\pdfoutput=1
\newcommand{\forcamera}[1]{}

\documentclass[11pt]{article}

\usepackage[table]{xcolor}
\newcommand{\NEW}{\rowcolor{gray!50}}

\usepackage{acl}

\usepackage{tgtermes}
\usepackage{latexsym}

\usepackage{todonotes}

\usepackage{caption}

\usepackage[T1]{fontenc}
\usepackage[T5]{fontenc}
\usepackage[utf8]{inputenc}

\usepackage{microtype}

\usepackage{makecell}

\usepackage{pifont}

\newcommand*\rot{\rotatebox{90}}

\usepackage{color}

\usepackage{float}
\usepackage{makecell}
\usepackage{minted}

\usepackage{pgfplots}
\pgfplotsset{compat=1.16}
\usepackage{tikz}
\usepackage{amsmath}

\usepackage{flafter} %

\newcommand{\caancora}[0]{\textsf{ca\_ancora}}%
\newcommand{\cspcedt}[0]{\textsf{cs\_pcedt}}%
\newcommand{\cspdt}[0]{\textsf{cs\_pdt}}%
\newcommand{\cspdtsc}[0]{\textsf{cs\_pdtsc}}%
\newcommand{\cuproiel}[0]{\textsf{cu\_proiel}}%
\newcommand{\deparcorfull}[0]{\textsf{de\_parcorfull}}%
\newcommand{\depotsdamcc}[0]{\textsf{de\_potsdam}}%
\newcommand{\enfantasycoref}[0]{\textsf{en\_fantasycoref}}%
\newcommand{\engum}[0]{\textsf{en\_gum}}%
\newcommand{\enlitbank}[0]{\textsf{en\_litbank}}%
\newcommand{\enparcorfull}[0]{\textsf{en\_parcorfull}}%
\newcommand{\esancora}[0]{\textsf{es\_ancora}}%
\newcommand{\francor}[0]{\textsf{fr\_ancor}}%
\newcommand{\frdemocrat}[0]{\textsf{fr\_democrat}}%
\newcommand{\frlitbankfr}[0]{\textsf{fr\_litbankfr}}%
\newcommand{\grcproiel}[0]{\textsf{grc\_proiel}}%
\newcommand{\hboptnk}[0]{\textsf{hbo\_ptnk}}%
\newcommand{\hihdtb}[0]{\textsf{hi\_hdtb}}%
\newcommand{\huszegedkoref}[0]{\textsf{hu\_szeged}}%
\newcommand{\hukorkor}[0]{\textsf{hu\_korkor}}%
\newcommand{\koecmt}[0]{\textsf{ko\_ecmt}}%
\newcommand{\lacoreflat}[0]{\textsf{la\_coreflat}}%
\newcommand{\ltlcc}[0]{\textsf{lt\_lcc}}%
\newcommand{\nlopenboek}[0]{\textsf{nl\_openboek}}%
\newcommand{\nobokmaalnarc}[0]{\textsf{no\_bokmaalnarc}}%
\newcommand{\nonynorsknarc}[0]{\textsf{no\_nynorsk\-narc}}%
\newcommand{\plpcc}[0]{\textsf{pl\_pcc}}%
\newcommand{\rurucor}[0]{\textsf{ru\_rucor}}%
\newcommand{\tritcc}[0]{\textsf{tr\_itcc}}%

\newcommand{\sys}[1]{\textsc{#1}}
\newcommand{\baseline}[0]{\sys{Baseline}}
\newcommand{\baselinegz}[0]{\sys{Baseline-GZ}}

\def\MC#1#2{\multicolumn{#1}{c}{#2}}

\title{Findings of the Fifth Shared Task on Multilingual Coreference Resolution: 
Expanding Datasets for Long-Range Entities}

\author{
Michal Novák$^1$,
Miloslav Konopík$^2$,
Anna Nedoluzhko$^1$,
Martin Popel$^1$, \\
\textbf{Ondřej Pražák$^2$,
Jakub Sido$^2$,
Milan Straka$^1$,
Zdeněk Žabokrtský$^1$,
Daniel Zeman$^1$} \\[2mm]
$^1$ Charles University, Faculty of Mathematics and Physics, \\ Institute of Formal and Applied Linguistics, Prague, Czechia \\
\texttt{\{mnovak,nedoluzko,popel,straka,zabokrtsky,zeman\}@ufal.mff.cuni.cz}\\[2mm]
$^2$
University of West Bohemia, Faculty of Applied Sciences, \\ Department of Computer Science and Engineering, Pilsen, Czechia \\
\texttt{\{konopik,ondfa,sidoj\}@kiv.zcu.cz}\\[2mm]
}

\usepackage{booktabs}
\newcommand{\ndatasets}[0]{27}
\newcommand{\nlanguages}[0]{19} %
\newcommand{\nsystems}[0]{ten} %
\newcommand{\nteams}[0]{eight} %

\usepackage{fancyhdr}
\fancypagestyle{officialbibref}{\fancyhf{}\fancyheadoffset[L,R]{50pt}\fancyhead[C]{This paper was published at \textbf{CODI-CRAC 2026} -- please cite the published version {\small\url{https://aclanthology.org/2026.codi-1.21}}.}\fancyfoot[C]{\normalsize\thepage}\addtolength{\topmargin}{-30pt}\addtolength{\headsep}{30pt}}
\begin{document}
\thispagestyle{officialbibref}
\pagenumbering{arabic}\pagestyle{plain}
\maketitle

\begin{abstract}

This paper describes the fifth edition of the Shared Task on Multilingual Coreference Resolution, held in conjunction with the CODI-CRAC 2026 workshop. Building on previous iterations, the task required participants to develop systems capable of mention identification and identity-based coreference clustering.

The 2026 edition specifically emphasizes long-range entities, defined as coreferential chains spanning significant distances, across many words and sentences.

The task expanded its linguistic scope by incorporating five new datasets and two additional languages. These additions leverage version 1.4 of CorefUD, a harmonized multilingual collection comprising \ndatasets{} datasets in \nlanguages{} languages.

In total, \nsystems{} systems participated, including four LLM-based approaches
(three fine-tuned models and one few-shot approach).
While traditional systems still maintained their lead,
LLMs demonstrated significant potential, suggesting they may soon challenge established approaches in future editions.
\end{abstract}

\section{Introduction}

Coreference is the phenomenon in which multiple expressions in a discourse refer to the same real-world entity. For example, in \emph{``Beethoven was a revolutionary artist. The German composer changed the course of music, and he continues to inspire musicians today.''}, the mentions \emph{``Beethoven''}, \emph{``the German composer''}, and \emph{``he''} all denote the same individual. The goal of \emph{coreference resolution} is to automatically identify such mentions and group them into entity clusters. In the multilingual setting, the task is complicated by the diversity of languages and their grammatical and discourse conventions.

This paper presents the shared task on multilingual coreference resolution based on the CorefUD collection. Building on previous editions, we aim to provide a common evaluation framework and a multilingual benchmark that supports fair comparison across modeling paradigms.

A particular focus of the most recent edition was the emergence of large language models (LLMs). While non-LLM approaches remained welcome in the \emph{Unconstrained track}, we introduced a dedicated \emph{LLM Track} to explicitly study the strengths and limitations of LLM-based coreference systems. The previous edition indicated that LLM-based submissions still lagged behind the best discriminative encoder-based systems; given the rapid pace of LLM research, we therefore repeated the LLM Track to reassess the current state of LLM-based coreference. It targets solutions that primarily rely on LLMs, including fine-tuning, in-context learning, prompt design, constrained decoding, and more complex agentic pipelines.

Beyond modeling, the shared task also addresses a long-standing limitation of coreference benchmarks. Widely used resources such as OntoNotes \citep{ontonotes5.0} predominantly consist of short-to-medium documents (often below 1k tokens), largely because coreference annotation becomes increasingly difficult and costly as context grows: annotators typically resolve references incrementally and must track entities across long spans. Consequently, several projects either partition long documents into smaller units (including OntoNotes, later re-merged in LongtoNotes \citep{shridhar-etal-2023-longtonotes}; similarly, PDTSC, newly added to CorefUD, limits its documents to 50 sentences \citep{mikulova-etal-2017-pdtsc}; see Section~\ref{sec:data-new}) or constrain the annotation scope (e.g., PROIEL datasets \citep{Haug2008CreatingAP} limit the window to 13 sentences).\footnote{See their Guidelines for annotation of givenness at \url{https://dev.syntacticus.org/proiel-givenness-annotation-v1.pdf}} Because of the latter, long documents do not necessarily imply long-range coreference. We therefore focus on the \emph{range} of an entity, defined as the number of words between the first and last mention heads.

Despite these annotation complexities, new datasets targeting long-range coreference---often derived from literary texts---have emerged, partly driven by increased interest in NLP for literary domains \citep{bamman-etal-2014-bayesian,jcls3924}. Examples include
English: LitBank \citep{Bamman20Litbank}, FantasyCoref \citep{han-etal-2021-fantasycoref}, and BookCoref \citep{martinelli-etal-2025-bookcoref};
Dutch: RiddleCoref \citep{vanCranenburgh2019dutchcoref} and OpenBoek \citep{Cranenburgh_vanNoord_2022};
Korean: KoCoNovel \citep{kim-etal-2024-koconovel};
German: GerDraCor-Coref \citep{pagel-reiter-2020-gerdracor};
French: LitBank-fr \citep{MelanieBecquet2024BookNLPfr} and its extension Long-LitBank-fr \citep{bourgois-poibeau-2025-elephant}.
In parallel, there is growing work on coreference modeling in long contexts \citep{toshniwal-etal-2020-learning,toshniwal-etal-2021-generalization,guo-etal-2023-dual,gupta-etal-2024-coreference,martinelli-etal-2024-maverick,martinelli-etal-2025-xcore,bourgois-poibeau-2025-elephant}.

Most existing approaches and datasets, however, have been evaluated only monolingually. This motivated us to incorporate selected long-range entity resources into CorefUD, enabling multilingual evaluation of methods designed for long contexts. Because annotating long texts is challenging, many long-document datasets restrict the set of annotated entity types (e.g., LitBank distinguishes only six types, and some corpora focus exclusively on characters, such as BookCoref or KoCoNovel). For this shared task, we therefore prioritize datasets with a broader range of annotated entities; while character-only datasets are highly appealing (e.g., due to their length or coverage of underrepresented languages), including them would have required substantial changes to the evaluation framework that were not feasible within the available time.

For these reasons, we extend CorefUD with English FantasyCoref, French LitBank-fr, and Dutch OpenBoek.

In addition, we add two further datasets, introducing Latin and increasing the number of speech-domain datasets. Finally, for the Unconstrained track, we also support input data in a simplified JSON format to lower the barrier to participation.

The shared task was hosted on Codabench;\footnote{LLM Track: \url{https://www.codabench.org/competitions/13198}; Unconstrained Track: \url{https://www.codabench.org/competitions/13197}} the competition was re-opened a few months after the evaluation phase to enable future evaluation on the same test set.

The remainder of the paper is structured as follows.
Section~\ref{sec:data} describes the data used in the shared task, while 
Section~\ref{sec:evaluation} presents the evaluation methodology.
Section~\ref{sec:systems} presents and compare baseline and participating systems.
We report and analyze the official results in Section~\ref{sec:results}, and conclude in Section~\ref{sec:conclusions}.
\section{Datasets}
\label{sec:data}

\begin{table*}[!htb]
  \begin{center}
    \resizebox{\textwidth}{!}{
\begin{tabular}{@{}l rrrr rrrr r rrrr @{}}\toprule
                              & \MC{4}{text size}                  & \MC{4}{entities}              &          & \MC{4}{mentions}              \\\cmidrule(lr){2-5}\cmidrule(lr){6-10}\cmidrule(lr){11-14}
document                      & \MC{4}{total number of}            &  total &per 1k &\MC{2}{length}& range    &  total &per 1k &\MC{2}{length}\\\cmidrule(lr){2-5}\cmidrule(lr){8-9}\cmidrule(lr){13-14}
                              &  docs &  sents &    words &empty n.&  count & words &   max &  avg & p95      &  count & words &   max &  avg \\\midrule
Ancient\_Greek-PROIEL         &    19 &  6,475 &   64,111 &  6,283 &  3,215 &    50 &   332 &  6.6 &      290 & 21,354 &   333 &    52 &  1.7 \\
Ancient\_Hebrew-PTNK          &    40 &  1,161 &   28,485 &      0 &    870 &    31 &   102 &  7.2 &      744 &  6,247 &   219 &    22 &  1.5 \\
Catalan-AnCora                & 1,298 & 13,613 &  429,313 &  6,377 & 17,558 &    41 &   101 &  3.6 &      538 & 62,417 &   145 &   141 &  4.8 \\
Czech-PCEDT                   & 2,312 & 49,208 &1,155,755 & 35,315 & 49,036 &    42 &   236 &  3.4 &      663 &167,418 &   145 &    79 &  3.6 \\
Czech-PDT                     & 3,165 & 49,419 &  834,707 & 21,092 & 46,460 &    56 &   173 &  3.3 &      430 &154,437 &   185 &    99 &  3.1 \\
\NEW Czech-PDTSC              & 1,553 & 73,802 &  743,963 & 62,130 & 44,940 &    60 &    81 &  4.1 &      400 &185,512 &   249 &    57 &  1.6 \\
\NEW Dutch-OpenBoek           &     9 &  5,704 &  103,518 &      0 &  2,287 &    22 &   780 &  7.4 &\bf 8,708 & 16,995 &   164 &    20 &  1.5 \\
\NEW English-FantasyCoref     &   211 & 12,126 &  340,545 &      0 &  5,829 &    17 &   339 &  9.8 &\bf 2,087 & 56,963 &   167 &    50 &  1.9 \\
English-GUM                   &   237 & 13,263 &  233,926 &    121 &  9,266 &    40 &   131 &  4.4 &      899 & 40,918 &   175 &    95 &  2.6 \\
English-LitBank               &   100 &  8,560 &  210,530 &      0 &  2,164 &    10 &   261 & 10.8 &\bf 1,939 & 23,340 &   111 &   129 &  1.6 \\
English-ParCorFull            &    19 &    543 &   10,798 &      0 &    188 &    17 &    38 &  4.4 &      556 &    835 &    77 &    37 &  2.1 \\
French-ANCOR                  &   455 & 31,761 &  454,577 &      0 & 13,204 &    29 &   103 &  4.3 &\bf 2,073 & 56,459 &   124 &    17 &  1.9 \\
French-Democrat               &   126 & 13,057 &  284,883 &      0 &  7,162 &    25 &   895 &  6.5 &\bf 7,440 & 46,487 &   163 &    71 &  1.7 \\
\NEW French-LitBankFr         &    29 & 11,312 &  276,985 &      0 &  2,023 &     7 &   934 & 16.9 &\bf11,583 & 34,228 &   124 &    22 &  1.5 \\
German-ParCorFull             &    19 &    543 &   10,602 &      0 &    243 &    23 &    43 &  3.7 &      487 &    896 &    85 &    30 &  2.0 \\
German-PotsdamCC              &   176 &  2,238 &   33,222 &      0 &    880 &    26 &    15 &  2.9 &      165 &  2,519 &    76 &    34 &  2.6 \\
Hindi-HDTB                    &   271 &  3,479 &   76,282 &      0 &  3,148 &    41 &    36 &  3.8 &      358 & 12,082 &   158 &    43 &  1.8 \\
Hungarian-KorKor              &    94 &  1,351 &   24,568 &  1,569 &  1,122 &    46 &    41 &  3.6 &      244 &  4,091 &   167 &    42 &  2.2 \\
Hungarian-SzegedKoref         &   400 &  8,820 &  123,968 &  4,857 &  4,769 &    38 &    36 &  3.2 &      327 & 15,165 &   122 &    36 &  1.6 \\
Korean-ECMT                   & 1,470 & 30,784 &  482,986 &      0 & 16,536 &    34 &    55 &  3.4 &      448 & 56,538 &   117 &    12 &  1.3 \\
\NEW Latin-CorefLat           &    16 &  1,741 &   25,965 &      0 &    734 &    28 &    14 &  2.8 &       82 &  2,075 &    80 &    70 &  2.5 \\
Lithuanian-LCC                &   100 &  1,714 &   37,014 &      0 &  1,087 &    29 &    23 &  4.0 &      385 &  4,337 &   117 &    19 &  1.5 \\
Norwegian-BokmaalNARC         &   346 & 15,742 &  245,515 &      0 &  5,658 &    23 &   298 &  4.7 &    1,042 & 26,611 &   108 &    51 &  1.9 \\
Norwegian-NynorskNARC         &   394 & 12,481 &  206,660 &      0 &  5,079 &    25 &    84 &  4.3 &      852 & 21,847 &   106 &    57 &  2.1 \\
Old\_Church\_Slavonic-PROIEL  &    26 &  6,832 &   61,759 &  6,289 &  3,396 &    55 &   134 &  6.5 &      279 & 22,116 &   358 &    52 &  1.5 \\
Polish-PCC                    & 1,828 & 35,874 &  538,885 & 18,615 & 22,143 &    41 &   135 &  3.7 &      276 & 82,706 &   153 &   108 &  1.9 \\
Russian-RuCor                 &   181 &  9,035 &  156,636 &      0 &  3,515 &    22 &   141 &  4.6 &    1,064 & 16,193 &   103 &    18 &  1.7 \\
Spanish-AnCora                & 1,356 & 14,159 &  458,418 &  8,112 & 19,445 &    42 &   110 &  3.6 &      530 & 70,663 &   154 &   101 &  4.8 \\
Turkish-ITCC                  &    24 &  4,732 &   55,358 & 11,584 &  4,019 &    73 &   369 &  5.4 &    1,032 & 21,569 &   390 &    31 &  1.1 \\
\bottomrule\end{tabular}
}
    \caption{CorefUD~1.4 data sizes in terms of the total number of documents, sentences,
      words (i.e.\ non-empty nodes), empty nodes (empty words),
      coreference entities
      (total count, relative count per 1000 words, average and maximal length in number of mentions and 95th percentile of range, i.e. distance from the first to the last mention in words)
      and coreference mentions
      (total count, relative count per 1000 words, average and maximal length in number of words).
      All the counts are excluding singletons and for the concatenation of train+dev+test.
      Train/dev/test splits of these datasets roughly follow the 8/1/1 ratio,
      but for the shared task we used reduced versions: mini-dev and mini-test.
      Newly added datasets have a gray background
      and values of p95 range$>$1500 are in bold.
      }
    \label{tab:sizes}
  \end{center}
\end{table*}

As in previous years, the shared task uses training and evaluation data from
the public part of the CorefUD collection
\cite{corefud2022lrec},\footnote{\url{https://ufal.mff.cuni.cz/corefud}} now
available in its latest release
(1.4).\footnote{\url{https://hdl.handle.net/11234/1-6108}} The public edition
of CorefUD~1.4 contains 29 datasets%
\footnote{For the shared task, we used only 27 of them (see
Section~\ref{sec:data-preproc}).} covering \nlanguages{} languages. Compared
to CorefUD~1.3, used last year \cite{oursharedtask2025}, the new release adds
five datasets and two languages. Several of the additions emphasize
long-document collections (primarily literary texts), which can challenge
systems that are often optimized for short-range coreference.
In addition to English LitBank and French Democrat\&ANCOR
from previous editions, long-entity\footnote{
 The entity length can be defined in various ways, e.g., by the number of mentions.
 However, here we focus rather on the \emph{range} of the entity,
  i.e., the distance from the first to the last mention of a given entity in words.
 We report this entity range in Table~\ref{tab:sizes} as the 95th percentile for each dataset,
  i.e., the longest range after excluding top 5~\% (possible outliers) and also excluding singletons.
 When the p95 range is higher than 1500, we consider a given dataset a long-entity dataset.
} 
data are now represented by French LitBankFr, Dutch OpenBoek, and English FantasyCoref.
With OpenBoek, we also add Dutch to the language set.
The set of languages is further expanded with Latin via CorefLat, and we broaden
the coverage of spoken language by including Czech PDTSC. Overall, the datasets
span diverse domains, including news, fiction, Bible texts, and Wikipedia
articles. Table~\ref{tab:sizes} provides an overview of the datasets and
their sizes; see Appendix~\ref{sec:data-references} for references to the
individual datasets.

A key aim of the CorefUD project is to stimulate research on coreference resolution beyond English, especially for languages that exhibit zero anaphora.
Zero anaphora (\emph{zero mentions}) arises when a referent (e.g., a subject or object) is recoverable from context but not overtly realized. This is typical of pro-drop languages, where verbal morphology often provides sufficient cues to reconstruct the omitted pronoun.
In CorefUD, such cases are encoded as \emph{empty nodes} inserted into Universal Dependencies (UD) trees, allowing them to participate in coreference chains alongside overt mentions.
Compared to the previous version, the set of datasets with zero mentions has been extended by introducing Czech PDTSC.

Our shared task is restricted to identity coreference. The CorefUD datasets, however, may also annotate other relations (e.g., bridging), and some corpora additionally mark phenomena such as event or abstract anaphora, while others do not. Since CorefUD is not fully harmonized with respect to annotation guidelines, the exact interpretation of these anaphoric labels can differ slightly across corpora. In converting resources to the CorefUD format, we aim to extract identity coreference\footnote{We are aware that complete isolation is not possible due to near-identity relations; see \citet{recasensNearIdentity2010}.} while otherwise preserving the original annotation as much as possible.

\subsection{New Resources}
\label{sec:data-new}

\paragraph{English FantasyCoref} English FantasyCoref \citep[\enfantasycoref{};][]{han-etal-2021-fantasycoref} is a conversion of the FantasyCoref corpus consisting of English fantasy texts. It provides full entity level coreference annotation and follows an entity cluster model. Annotations are made from an omniscient writer’s point of view, extending OntoNotes guidelines to account for literary specific phenomena such as asymmetry of knowledge, transformations or status changes of entities, foretold or wished entities that later materialize, and extensive lexical or metaphorical variation. The corpus is larger and structurally more complex than \enlitbank, with longer documents and longer range coreference chains.

\paragraph{French LitBank} \citep[\frlitbankfr;][]{jcls3924} is a corpus of 19th--20th century French literary prose. The texts consist mainly of novel excerpts originating from the French-Democrat corpus (\frdemocrat), and are annotated with literary entities and coreference relations. Entity annotation follows principles adapted from the English-LitBank corpus and is limited to selected categories relevant for literary analysis (e.g., persons, locations, facilities, geopolitical entities, vehicles, and time expressions). Coreference annotation targets these entities and is designed to support long-distance reference tracking typical of literary texts.

\paragraph{Dutch OpenBoek} \citep[\nlopenboek;][]{Cranenburgh_vanNoord_2022} is a corpus of Dutch literary prose from the 19th and early 20th centuries, consisting of long fragments (typically over 10\,000 tokens) from public-domain novels. The corpus is annotated for nominal and pronominal coreference, including singletons, using a single identity-based coreference category.
\paragraph{Czech PDTSC} \citep[\cspdtsc;][]{pdtsc}, i.e.
Prague Dependency Treebank of Spoken Czech, contains
transcriptions of spontaneous dialogues from the projects Malach (testimonies
of Shoah survivors) and Companions (a virtual companion chatbot for lonely
people). In both cases, the dialogues are asymmetric, with one party speaking
most of the time and the other party inserting short comments and questions.
The speech is transcribed using standard Czech orthography, including
punctuation.
For technical reasons, the conversations are split into documents of 50 sentences, and coreference is only annotated inside each document.

\paragraph{Latin CorefLat} \citep[\lacoreflat;][]{delfino2024building}
is a converted version of CorefLat, a coreference-annotated corpus of Latin. It comprises four literary texts: two prose works (Book~I of \textit{De bello Gallico} by Gaius Julius Caesar; Book~I of \textit{Confessiones} by Augustine of Hippo) and two theatrical plays (\textit{Curculio} by Titus Maccius Plautus; \textit{Medea} by Lucius Annaeus Seneca).

Unlike other CorefUD datasets, CorefLat spans several centuries. The texts range from Curculio (2nd century BCE), the oldest, to Confessiones (4th century CE), the newest, which may already allow for the study of coreference from a diachronic perspective. For the purposes of the CorefUD collection, however, we adopted a train/dev/test split that preserves the overall proportion of each text across sections: roughly the first 10~\% of each of the four documents is allocated to the test set, another 10~\% to the development set, and the remaining 80~\% to the training set.

The CorefLat dataset appears to underperform compared to the results reported for other CorefUD datasets later in this paper (see Table~\ref{tab:all-langs}), most notably for the baseline system, which achieves only 6.8 on the primary metric. In a post--shared task investigation, we identified a flaw in the conversion pipeline. In addition to the link-based coreference representation, which was correctly imported into CorefUD~1.4, CorefLat also uses a cluster-based representation for certain groups of mentions; this representation was not handled by the converter. After extending the converter to combine both representation types, a preliminary experiment with the baseline system reached a CoNLL F1 score of 37.
We therefore expect that releasing an improved version of the dataset will also substantially improve the performance of other systems. Nevertheless, CorefLat may still pose specific challenges for coreference resolution because of its diachronic character and genre diversity.

\bigskip
The original \enfantasycoref, \frlitbankfr, \nlopenboek, and \lacoreflat{} corpora were not annotated in the UD framework; therefore, morpho-syntactic annotation in the CorefUD version is produced by UDPipe~2.

\subsection{Updated Resources}
\label{sec:data-update}

The English GUM corpus (\engum) contains the same sentences as in CorefUD~1.3, but with minor corrections in the annotations from UD~2.17.

All three Czech datasets (PDT, PCEDT, and PDTSC) are subsets of the Prague
Dependency Treebank-Consolidated, PDT-C~2.0, converted to UD \cite{udpdtc}.
While PDTSC is new in CorefUD, the other two were included in previous
versions of CorefUD. Changes since the last version are in morphosyntactic
annotation and its PDT-to-UD conversion. Coreference annotation is unchanged,
except for a few corrections.

\paragraph{Updated morphosyntax prediction}
For datasets without manual morphosyntactic annotation, we re-predicted UD relations, tags, and features using newer UDPipe models based on UD release 2.17 (previously 2.15).
Czech PCEDT was excluded from this subset because the latest version now includes manual morphosyntactic annotation.

\subsection{Data for the Shared Task}
\label{sec:data-preproc}

Relative to the public CorefUD~1.4 release, the dataset distributed to shared task participants was subject to minor adjustments.%
\footnote{
Both the shared task data and submissions are available at \url{http://hdl.handle.net/11234/1-6160}.
}

\paragraph{Data reduction}
As in last year's shared task, the English and German ParCorFull datasets were excluded due to their very small size and high variance, which disproportionately affected macro-averaged results. Moreover, the development and test sets were reduced to \emph{mini-dev} and \emph{mini-test} sets by capping each split at 25k words via random sampling of complete documents to lower evaluation costs (details in \citet{oursharedtask2025}).

New this year, we do not apply the 25k-word cap to the literary datasets with long documents, as capping would leave too few documents per split and reduce diversity across documents. In practice, this exception affects only \enfantasycoref{} and \frlitbankfr{}.

\begin{figure*}
    \centering
    \includegraphics[width=\linewidth]{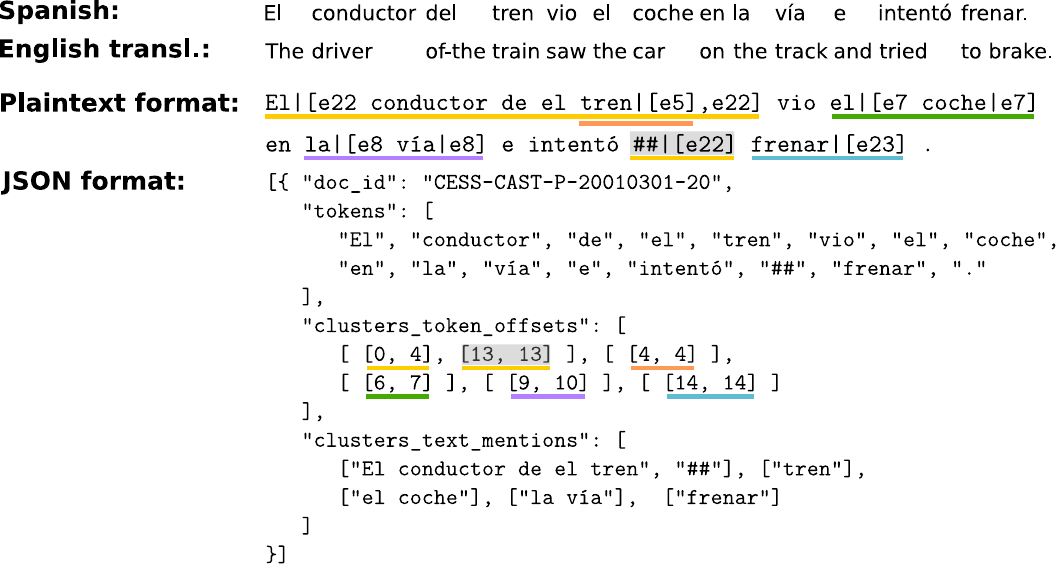}
    \caption{Serialization of a Spanish example sentence from \esancora{} to our plaintext and JSON format. For clarity, mention spans are highlighted by colored underlining, where two coreferential entities share the same color. A zero mention labeled on an empty node is grayed. Note that multi-word tokens are split in the plaintext format into syntactic words (e.g., the Spanish \emph{``del''} appears as \emph{``de el''})}
    \label{fig:plaintext-example}
\end{figure*}

\paragraph{Plaintext format}
For the LLM track, we provide a conversion of the data into a simplified plaintext representation, together with both the conversion utility and the already converted dataset files.

In this plaintext format (see Figure~\ref{fig:plaintext-example}), each document is stored as a single line of text, with tokens separated by spaces. Coreference information is attached to each token following the `\texttt{|}' symbol. Mentions, including singletons, are specified by their span boundaries, indicated using opening and closing square brackets combined with an entity identifier. Empty nodes are marked with the prefix '\texttt{\#\#}'. Since empty nodes are defined by syntactic structure rather than linear token order, they are inserted directly after their syntactic parent. This representation does not capture the type of dependency relation to the parent, and therefore cannot differentiate between multiple empty nodes sharing the same parent (see Section~\ref{sec:evaluation}). Although this restriction may have a minor influence on evaluation, we consider it a reasonable compromise to maintain the format’s simplicity.

Compared to CoNLL-U, the plaintext format is deliberately less expressive and does not retain enough information for certain evaluation metrics (for example, head match relies on identifying mention heads via syntactic trees). To address this, we provide a reverse conversion tool that reconstructs CoNLL-U files from the plaintext annotations, along with an output cleaning utility.%
\footnote{The conversion tool and cleaner are distributed as a unified application/Python library on GitHub: \url{https://github.com/ondfa/text2text-coref}}

The cleaner is designed to fix common issues in LLM-generated outputs, such as malformed annotation structures (e.g., missing mention boundaries) or alterations to the token sequence. It first enforces correct opening and closing of all mentions, and then aligns the output text with the original input using word-level edit distance. During this alignment, empty nodes are excluded, as systems are expected to generate them independently. After achieving an exact match between token sequences, the annotations can be reliably projected back onto the original CoNLL-U data.

\paragraph{JSON format}

In order to facilitate the development of participants' systems
in the unconstrained track,
we also provide data in a much simpler JSON format (see Figure~\ref{fig:plaintext-example}). We also offer tools for exporting coreference annotations from CoNLL-U into JSON and importing them back. The structure of the JSON format used in the Shared task is inspired by the structure of the output provided by the Maverick Coref package.\footnote{\url{https://github.com/SapienzaNLP/maverick-coref}}

 A file in JSON format is a list of documents, where each document is represented as a JSON object with four fields:
\begin{itemize}
    \item \textit{doc\_id}: the document ID
    \item \textit{tokens}: a list of tokens in the document.
    \item \textit{clusters\_token\_offsets}: a list of coreference clusters, where each cluster is represented as a list of mentions, and each mention is represented as a list of two integers indicating the start and end token offsets (inclusive) of the mention in the tokens list.
    \item \textit{clusters\_text\_mentions}: a list of coreference clusters, where each cluster is represented as a list of textual mentions corresponding to the mentions in clusters\_token\_offsets.
\end{itemize}

Conversion is very similar to the plaintext format. Empty nodes are represented exactly the same way; prefixed with `\textit{\#\#}', placed right after their syntactic head.

\paragraph{Data variants and starting points}
For each track, we release two versions of the data: \emph{gold data} (for training and evaluation against gold annotations) and \emph{input data} (the version intended to be processed by participants' systems and submitted).
For the Unconstrained track, we additionally provide three alternative starting points that differ in how much of the pipeline is already pre-filled with baseline predictions.

\emph{Gold data} contains gold-standard coreference and gold empty nodes, and is meant for training and evaluation.
It follows the CorefUD~1.4 release and preserves manually annotated morpho-syntactic features whenever they are available in the original dataset.
All CoNLL-U fields are kept as in the source data, including the forms of empty nodes.\footnote{Earlier data versions removed them to match the output of the baseline empty-node predictor. This is no longer necessary, since the baseline now predicts all CoNLL-U fields for empty nodes (see Section~\ref{sec:baseline}).}
The gold train and mini-dev sets were available for download; the gold mini-test set remained private and was used internally in Codabench for evaluation.

\emph{Input data} is derived from the gold data by applying preprocessing that mimics a setting without manual linguistic annotation.
This preprocessing was performed only for the mini-dev and mini-test sets.
Specifically, we replaced the original morpho-syntactic features with the outputs of UD~2.17 models across all datasets (including those that originally contained human-annotated features) and removed gold empty nodes and gold coreference annotations.
This yields the input data for the LLM track; for the Unconstrained track, participants could choose one of three \emph{starting points}: (1) \emph{Coreference and zeros from scratch} (no predictions of empty nodes or coreference; identical to the LLM-track input), (2) \emph{Coreference from scratch} (baseline empty-node predictions provided, no coreference), and (3) \emph{Refine the baseline} (baseline predictions of both empty nodes and coreference provided).

\section{Evaluation Metrics}
\label{sec:evaluation}

As in prior editions, the primary evaluation score is the CoNLL F$_1$ score, computed using head mention matching with singletons excluded, and a dependency-based procedure to match zero mentions.
Alongside the primary score, we report additional metrics to enable a broader comparison of submitted systems.

\paragraph{Official scorer}
Outputs of participating systems are scored with the CorefUD scorer,%
\footnote{\url{https://github.com/ufal/corefud-scorer}}
which has not been changed since the previous editions.
The scorer is based on the Universal Anaphora (UA) scorer 2.0 \citep{ua-scorer-2.0},
and inherits all functionality needed for the shared task, including implementations of standard coreference evaluation metrics.
Compared to the UA scorer, the CorefUD scorer additionally provides head matching and a dependency-based alignment method for zero mentions.

The scorer expects two CoNLL-U files: a gold file and a system prediction.
Because both our plaintext and JSON formats omit some information required for evaluation (e.g., mention heads), any output produced in these formats must be reconstructed into CoNLL-U prior to evaluation.

\paragraph{Mention matching}
Given the shortcomings of \emph{exact} and \emph{partial} mention matching (see \citet{oursharedtask2023} for details), we use \emph{head match} strategy for the primary metrics.
Under this criterion, a gold mention and a predicted mention are matched when they share the same head token.%
\footnote{In CorefUD, gold mention heads are derived from the dependency tree using the Udapi block \texttt{corefud.MoveHead}.}
Mention spans are otherwise disregarded, except when multiple mentions have an identical head, in which case spans are used to distinguish them.

Head matching, however, cannot be applied to empty nodes, which are necessary to represent zero anaphora.
A system's predicted counterpart of a gold zero may be absent, spurious, or placed at a different surface position in the sentence, while still expressing the same role.
We therefore employ the \emph{dependency-based method} \citep{oursharedtask2024}, which
 aligns predicted and gold zero mentions within each sentence by maximizing their overlap in the enhanced dependency representation.
Concretely, it solves a one-to-one matching problem in a weighted bipartite graph; each candidate pair is weighted by how well the predicted zero reproduces the gold zero’s dependencies.
Pairs that correctly recover both the parent and the dependency label are rewarded more, while the procedure still works reliably even when labels are missing.

\paragraph{Primary score}
Following common practice in coreference resolution, the primary metric is CoNLL F$_1$  \citep{CoNLL-MELA-score,pradhan-etal-2014-scoring}.
It is defined as the unweighted mean of the $F_1$ scores of three established measures: MUC \citep{MUC-score}, B$^3$ \citep{Bcubed-score}, and CEAF-e \citep{CEAF-score}.
These correspond to complementary viewpoints on coreference quality---link-based, mention-based, and entity-based, respectively.
To identify systems that perform consistently across datasets, we rank submissions by the macro-average of CoNLL F$_1$ over all mini-test sets in the shared task collection.%

\paragraph{Supplementary scores}
In addition to the main CoNLL F$_1$ score, we also report variants based on alternative mention matching strategies: partial match%
\footnote{Partial match served as the primary metric in the first edition of the shared task \citep{oursharedtask2022}.}
and exact match.
We further compute a head-match CoNLL score that includes all mentions, including singletons.

For a more detailed view, we provide the individual metrics that make up the CoNLL score (MUC, B$^3$, and CEAF), as well as other popular measures such as BLANC \citep{BLANC-score} and LEA \citep{LEA-score}.
Finally, we report the anaphor-decomposable score for zero anaphora and the mention overlap ratio (MOR), both introduced in \citet{oursharedtask2022}. We also introduce the mention-detection F-score on heads (MD-h).
Both mention-oriented scores assess mention detection quality independently of entity assignment. Whereas MOR assigns each mention pair a graded score based on full-span overlap, MD-h simplifies the comparison to a binary match between mention heads.

\section{Participating Systems}
\label{sec:systems}
\subsection{Baseline}
\label{sec:baseline}
As in the previous edition, we provide two baseline systems: one that predicts empty nodes as placeholders for zero anaphora and another for coreference resolution.
These baseline systems may be used or extended only by participants in the Unconstrained track.

\paragraph{Empty nodes prediction baseline}

As in the previous two years, empty node prediction is a part of the shared task. To
assist participants focusing solely on coreference resolution, we provide
a baseline system for empty node prediction.

This year, the empty node prediction baseline has been updated considerably.
In previous years, the baseline produced only information necessary for the
coreference resolution evaluation---for every empty node, its word order,
dependency parent, and dependency relation were predicted. This year, we extended
the system to predict all available empty node information, additionally providing
the form, lemma, UPOS, XPOS, and FEATS columns.

The baseline model architecture builds on the implementation from previous
years. Each input sentence is processed by
XLM-RoBERTa-large~\citep{conneau-etal-2020-unsupervised} to generate embeddings
for every word. Based on these embeddings, two candidate empty nodes are
predicted per word and processed through eight heads: (1)~a~binary
classification head to determine whether the empty node candidate will produce
an empty node, (2-7)~classification heads predicting the dependency relation,
form, lemma, UPOS, XPOS, and FEATS, and (8)~a self-attention-based word-order
prediction head to identify the insertion point for the empty node. A single
model is trained on a concatenation of all corpora containing empty nodes, with
sentences sampled proportionally to the square root of their respective corpus
sizes.

The source code of the system is available under an open-source license.\footnote{\url{https://github.com/ufal/crac2026_empty_nodes_baseline}}
The trained model is available on Hugging Face\footnote{\small\url{https://huggingface.co/ufal/crac2026_empty_nodes_baseline}} and via LINDAT/CLARIAH-CZ,\footnote{\small\url{https://hdl.handle.net/11234/1-6081}} and is automatically downloaded by the system.

Further information, including a detailed architecture description, training
hyperparameters, and an intrinsic evaluation of predicting all CoNLL-U columns for
all the datasets, is available in~\citet{straka-2026-corpipe}.

\paragraph{Coreference resolution baseline}
The coreference resolution baseline has remained unchanged over the past four years.%
\footnote{\url{https://github.com/ondfa/coref-multiling}}
It is built upon the multilingual end-to-end neural coreference resolution system by \citet{prazak-etal-2021-multilingual}, which extends the original end-to-end model introduced by \citet{lee-etal-2017-end}. The model examines all possible spans up to a predefined maximum length and directly predicts an antecedent for each span. As it does not include a separate mention detection stage, it is particularly suitable for datasets that do not annotate singleton mentions. The baseline uses the mBERT base model as its encoder \citep{DBLP:journals/corr/abs-1810-04805}.
In the remainder of this paper, we use \baseline{} to denote the combination of the two baseline systems, and \baselinegz{}, for coreference resolution baseline applied to gold empty nodes.

\subsection{System Submissions}
\label{sec:system-submissions}

This year, \nsystems{} systems were submitted to the shared task by \nteams{} teams:
UWB,\footnote{UWB = University of West Bohemia.}
LatticeNLP,\footnote{LatticeNLP refers to the system by Antoine Bourgois.}
Landcore,\footnote{Landcore refers to the system by Jan Pavelka.}
PortNLP,
AU-KBC,\footnote{AU-KBC = Anna University - K. B. Chandrasekhar Research Centre refers to the team of Sobha LalithaDevi, Pattabhi R K Rao and Vijay Sundar Ram.}
ÚFAL CorPipe,\footnote{ÚFAL CorPipe submitted three variants: CorPipeLarge, CorPipeXXL, and CorPipeEnsemble.}
Stanford NLP Group,\footnote{Stanford NLP Group is the creator of the Stanza package.}
and LEMN Lab.\footnote{LEMN Lab is the creator of DAggerCoref.}
For clarity, we distinguish the submissions to the LLM track with the `LLM-' prefix in the following text.

\paragraph{LLM-UWB (hejmanj)}
The UWB team fine-tunes large open-source models---primarily \texttt{google/gemma-3-27b-it}, alongside \texttt{gemma-3-12b-it} and \texttt{meta-llama/Llama-3.1-8B}---using QLoRA. Their approach modifies the plaintext format to tag solely the headword of a mention rather than its entire span. They deploy two distinct strategies: an end-to-end processing pipeline applied to most datasets, and a scrolling context window with overlap, which is specifically used on four datasets (namely Dutch-OpenBoek, French-Democrat, French-LitBankFr, Latin-CorefLat) to reduce inference costs. Training involves joint pretraining across all datasets followed by dataset-specific steps. Empty nodes are predicted directly within the generated text. In terms of generation capacity, the models process inputs up to 55,000 tokens and support immense output capacities of 131,072 tokens.

\paragraph{LLM-LatticeNLP (Antoine Bourgois)}
Based on the \texttt{gemma-3-27b-it} model, this system utilizes 4-bit quantization and LoRA adaptation (rank 64). The training relies on a two-stage supervised fine-tuning (SFT) strategy. First, a general adapter is trained across all datasets for one epoch to learn coreference phenomena and format. This is sequentially followed by one to four epochs of continual fine-tuning using a dataset-specific adapter to mitigate inter-dataset guideline discrepancies. The system uses a custom plaintext format relying exclusively on HTML/XML-like tags (e.g., \texttt{<ent8>} and \texttt{<zero2>}) inserted directly after the syntactic heads. During inference, documents are processed in batch queries of six sentences, augmented with an on-the-fly cleaning script to ensure valid contexts are passed to the subsequent generation windows.

\paragraph{LLM-Landcore (pavlk-mm)}
This system treats the shared task entirely as an applied in-context learning problem utilizing Anthropic's closed-source \texttt{Claude Haiku 4.5} model. No parameter updates or fine-tuning techniques are applied. The system transforms the input to an XML-like tagging scheme (e.g., \texttt{<e1>I</e1> sleep}). The constructed prompts are intricately detailed, describing linguistic observations and annotation suggestions alongside three corpus-specific few-shot examples comprising continuous chunks of up to 500 words (with the exception of \texttt{nl\_openbook}, which uses one example due to length). During inference, generations are requested at a temperature of 0. The system does not model or predict empty nodes.

\paragraph{LLM-PortNLP (portnlp)}
This system fine-tunes \texttt{Qwen-3-14B} using QLoRA with 4-bit NF4 quantization. Rather than processing full long documents at once, texts are split at sentence boundaries into 500--700 character chunks. The model resolves each chunk independently, relying computationally on a bounded context from prior generations alongside a novel ``scored entity registry.'' This registry handles up to 30 active entities mathematically managed via a frequency-recency decay formula. Training data included three language-specific few-shot examples, zero anaphora hints, and entity ID permutation augmentation to prevent the model from memorizing fixed entity IDs. To mitigate the dominance of resource-heavy languages (e.g., Czech) during training, temperature-scaled dataset sampling was implemented.

\paragraph{AUKBC-MULCRF (lalithadevi)}
As a participant in the Unconstrained track, this non-neural submission utilizes Conditional Random Fields (CRFs) to construct a multilingual coreference model from scratch. The system functions through a two-step pipeline. First, mentions are identified using UPOS tags, encompassing nouns, proper nouns, and pronouns. Subsequently, coreference chains are consolidated via noun-noun coreference and anaphora resolution methods that rely heavily on heuristic features (like edit distance) and syntactical cues (root words, dependencies, and PNG features). The approach completely omits the modeling of empty nodes.

\paragraph{CorPipeLarge (corpipe-sing-lg)}
This system represents an upgraded variant of their base CorPipe architecture from 2025, now pivoting to the \texttt{google/mt5-large} model comprising 580M parameters. Leveraging a word-level tagging scheme that supports overlapping clusters, the system dynamically predicts mentions and natively resolves coreference links directly via antecedent probabilities. Empty nodes are handled via integration with the shared task's provided baseline system. The system employs a purely multilingual approach, utilizing a batch consisting of 8 sentences, learning over 15 epochs without dataset-specific indicators.

\paragraph{CorPipeXXL (corpipe-single)}
Following the same architectural principles as CorPipeLarge, CorPipeXXL vastly scales the backbone model to \texttt{google/umt5-xxl}, harboring over 6B trainable parameters. To circumvent computational limitations and fit within standard memory constraints, smaller batches (6 sentences) and a reduced learning rate are employed. Additionally, rather than utilizing the exact published empty node baseline, the authors trained a slightly improved standalone baseline model using the provided implementation to inform their inputs.

\paragraph{CorPipeEnsemble (corpipe-ensemb)}
This submission represents a heavyweight amalgamation consisting of an ensemble of seven identically structured CorPipeXXL (\texttt{umT5-xxl}) models. Consequently, the holistic multi-model pipeline contains roughly 42 billion parameters. At inference time, performed expansively across seven parallel GPUs, the predictions are synthesized by averaging the word-level mention-encoding tag probabilities before subsequently averaging the decoded coreference link likelihoods across all constituent models.

\paragraph{Stanza (stanza-coref)}
Operating internally on the Stanza framework, the architecture functions as a head-joining efficient word-level coreference approach. Initially, the core model identifies mentions linked by
textual head words, whereupon semantic spans are resolved locally through an integrated Convolutional Neural Network (CNN). Representations are initialized using frozen \texttt{XLM-RoBERTa-large} embeddings conjoined with a LoRA adapter (34M trainable parameters), continuously sliding over the textual documents utilizing 512-token-wide windows. Empty nodes are acquired externally via the official task baseline.

\paragraph{DAggerCoref (thmorton)}
Submitted by the LEMN Lab to the Unconstrained track, DAggerCoref is a three-stage cascade built on \texttt{XLM-RoBERTa-large}. Stage 1 is a binary gap classifier that inserts learned \texttt{[ZERO]} markers at syntactically plausible empty-pronoun positions using an Otsu-adaptive threshold. Stage 2 is a token-level mention head classifier trained with focal loss. Stage 3 is a coarse-to-fine antecedent scorer using head-only mention representations. A central contribution is training Stage 3 with DAgger: after fitting on gold mentions, the system fine-tunes on a 50/50 mix of gold and predicted mention sequences to expose the scorer to test-time errors. The system is trained multilingually across all datasets without per-language tuning, predicting empty nodes entirely from scratch rather than relying on the provided baseline.

\begin{table*}[!t]
\centering
\begin{tabular}{l l l}
\hline
\textbf{Name} & \textbf{Track} & \textbf{Techniques} \\
\hline
LLM-UWB            & LLM       & FT, QLoRA, joint/scrolling window \\
LLM-LatticeNLP     & LLM       & FT, prompt tuning, QLoRA, quant., continual FT \\
LLM-Landcore       & LLM       & ICL, few-shot prompting \\
LLM-PortNLP        & LLM       & FT, QLoRA, quant., entity registry tracking \\
\hline
AUKBC-MULCRF       & Unconstr. & CRF, heuristic features pipeline \\
CorPipeLarge       & Unconstr. & FT joint mention/link \\
CorPipeXXL         & Unconstr. & FT joint mention/link \\
CorPipeEnsemble    & Unconstr. & FT + ensemble \\
Stanza             & Unconstr. & FT + LoRA, head-joining CNN \\
DAggerCoref        & Unconstr. & 3-stage cascade, DAgger FT, gap classifier \\
\hline
\end{tabular}
\caption{System names, task tracks, and main techniques.}
\label{tab:sys-tech}
\end{table*}

\begin{table*}[!t]
\centering
\begin{tabular}{l l r r r}
\hline
\textbf{Name} & \textbf{Model} & \textbf{Input ctx. len.} & \textbf{Output tok. len.} & \textbf{\#Params} \\
\hline
LLM-UWB            & \small\makecell[l]{Gemma-3-27B/-12B,\\ Llama-3.1-8B}  & 55,000      & 131,072    & 488 M (LoRA) \\
LLM-LatticeNLP     & gemma-3-27b-it               & 3,072       & 536        & 488 M (LoRA) \\
LLM-Landcore       & Claude Haiku 4.5             & 200,000     & 64,000     & —            \\
LLM-PortNLP        & Qwen-3-14B                   & 4,096       & 1,024      & 128.5 M (LoRA) \\
\hline
AUKBC-MULCRF       & CRF pipeline                   & —           & —          & —            \\
CorPipeLarge       & mt5-large                      & 512/2,560   & —          & 580 M        \\
CorPipeXXL         & umt5-xxl                       & 512/2,560   & —          & 6 B          \\
CorPipeEnsemble    & umt5-xxl ($\times 7$)          & 512/2,560   & —          & 42 B         \\
Stanza             & XLM-RoBERTa-large              & 512         & —          & \small\makecell[l]{34M + \\ 548M frozen}  \\
DAggerCoref        & XLM-RoBERTa-large              & 256/512     & —          & $\sim$582 M  \\
\hline
\end{tabular}
\caption{Models: model name, maximum input context length, maximum new tokens generated, and model sizes. Trainable LoRA parameter sizes are noted explicitly where applicable.}
\label{tab:context-size}
\end{table*}

\begin{table*}[!t]
\centering
\begin{tabular}{l l r r l}
\hline
\textbf{Name} & \textbf{Empty nodes} & \textbf{Batch size}    & \textbf{Grad ups} & \textbf{Tuned h-params} \\
\hline
LLM-UWB             & predicted    & 1--2           & $\sim$171 k      & model choice, window size   \\
LLM-LatticeNLP      & predicted    & 32             & $\sim$6 k        & plaintext type, epochs      \\
LLM-Landcore        & predicted    & few-shot       & 0                & —                           \\
LLM-PortNLP         & predicted    & 16             & $\sim$10.5 k     & none                        \\
\hline
AUKBC-MULCRF    & baseline           & NA             & NA               & —                           \\
CorPipeLarge    & baseline           & 8 sentences    & 150 k            & learning rate               \\
CorPipeXXL      & improved baseline  & 6 sentences    & 150 k            & learning rate               \\
CorPipeEnsemble & improved baseline  & 6 sentences    & 150 k $\times$ 7 & learning rate               \\
Stanza          & baseline           & 10 $\cdot$ 512-tokens & 255 k            & \small\makecell[l]{learning rate, warmup,\\LoRA params, \dots} \\
DAggerCoref     & predicted (scratch)& 4--32          & $\sim$115 k      & \small\makecell[l]{loss type, thresholds,\\DAgger mix, LRs, \dots} \\
\hline
\end{tabular}
\caption{Training configuration: empty-node handling, batch sizes, total gradient updates, and tuned hyperparameters. Stanza uses a batch size comprised of 10 windows (each 512-tokens long).}
\label{tab:train-config}
\end{table*}

\subsection{System Comparison}\label{sec:system-comparison}
\paragraph{Overview of tables}
Tables~\ref{tab:sys-tech}--\ref{tab:train-config} provide a comprehensive comparison of all submissions. Table~\ref{tab:sys-tech} enumerates each system’s shared-task track, primary pretrained architecture, and core methodological mechanisms (e.g., parameter efficient fine-tuning, in-context learning, entity registries, head-joining). Table~\ref{tab:context-size} details respective computational characteristics, mapping out each inference configuration’s input size, maximum generation cap, and total trainable parameter count. Finally, Table~\ref{tab:train-config} outlines specialized training configurations, documenting empty node conventions, batching heuristics, and total gradient updates.
While all teams leverage the official CoNLL-U dataset infrastructure, explicit divisions emerge concerning foundational text manipulation, context capacities, LLM application paradigms, and dataset-specific domain adaptations.

\paragraph{Modeling paradigms}
The task yielded four diverse contributions to the LLM track and six dedicated to the Unconstrained track. All LLM approaches opted to map the coreference resolution directly into a generative task. No agent or iterative approaches were involved. Submissions like LLM-UWB, LLM-LatticeNLP, and LLM-PortNLP employ supervised fine-tuning of selected large language models (Gemma-3, Qwen-3, Llama-3.1) utilizing quantized variants of the LoRA framework. The models were trained to generate either customized XML-like markup or target plaintext datasets. LLM-Landcore demonstrated an entirely training-free variation of generation leveraging well-engineered few-shot in-context prompts directed to Anthropic's proprietary Claude models.
Meanwhile, architectures in the Unconstrained track typically maintain traditional machine learning pipelines bridging mention detection functionality with sequential pair scoring algorithms. Teams predominantly utilize sequence-to-sequence encoder-decoder grids tailored explicitly to classification properties (CorPipe natively deploying \texttt{mT5}/\texttt{umT5}) or rely strictly on embedded representation analysis via CNNs tracking dense span overlaps (Stanza). Similarly leveraging \texttt{XLM-RoBERTa-large}, DAggerCoref introduces a three-stage cascade (gap classification, mention head classification, and antecedent scoring), uniquely employing DAgger fine-tuning to train on a mix of gold and predicted mentions. Amidst these neural approaches, one distinct outlier is AUKBC-MULCRF, electing to deploy highly specialized non-neural Conditional Random Fields (CRFs).

\paragraph{Context capacity and model scale}
Exploiting inherent advancements within foundational models, the contextual ceiling across systems drastically scaled compared to preceding iterations of the task. LLM-Landcore comfortably navigates prompts reaching limits of 200,000 subwords, whilst LLM-UWB optimized models to parse document scales generating effectively 131,072 target tokens. To offset computational overhead on larger language paradigms, system heuristics actively engineered clever solutions, such as LLM-PortNLP utilizing segmented bounding thresholds (500--700 characters) conjoined iteratively with a decay-driven scored entity registry. Machine learning approaches inside the Unconstrained boundaries still largely hinge on sequentially confined subword limits (512 blocks traversing to 2,560 inferences within the generalized CorPipe structures, or 256 to 512 subword windows in the case of DAggerCoref). Overall model parametrization varied drastically globally; the heaviest iteration utilized 42B cumulative variables (CorPipeEnsemble), dropping as far below as 34M trainable elements inside localized Stanza adapters, with mid-sized models like DAggerCoref sitting around 582M parameters.

\paragraph{Data usage and sampling considerations}
Universally adhering to competition constraints, external data was circumvented. However, internal sampling routines encountered notable shifts. Systems actively acknowledged the dataset disparities; for example, LLM-PortNLP intentionally executed temperature-scaled language sampling arrays, mathematically restricting the systemic Czech language distribution from heavily dominating cross-linguistic training batches. Likewise, randomization heuristics such as dataset ID permutation augmentation were utilized to ensure neural focus remained heavily fastened to linguistic features rather than entity ID integer memorization. Another novel approach to training data manipulation was introduced by DAggerCoref, which employs DAgger fine-tuning by explicitly mixing 50\% gold and 50\% predicted mention sequences during training to expose the model to realistic inference-time errors.

\paragraph{Empty node handling}
In approaches towards implicit zero-anaphora variables, system configurations split generally along paradigm constraints. For strictly generative networks resolving empty nodes natively (LLM-UWB, LLM-LatticeNLP, LLM-PortNLP, LLM-Landcore), resolutions happen end-to-end within internal text generations relying on distinct formatting rules (e.g. \texttt{<zero2>} injection mapping inline XML boundaries). %
Within the Unconstrained boundaries, baseline incorporation fundamentally ruled the architectures; AUKBC-MULCRF, CorPipeLarge, and Stanza strictly utilized predictions derived directly via the standard task baseline parameters, while CorPipeXXL and CorPipeEnsemble utilized a refined localized train-out of the same baseline framework. In contrast, DAggerCoref stands out in the Unconstrained track by discarding the baseline entirely, instead predicting empty nodes from scratch via a dedicated gap classifier that inserts learned \texttt{[ZERO]} sentinel tokens based on dependency information.

\paragraph{Language/treebank specialization and ensembling}
The architectural dichotomy balancing monolithic global models and dataset-specified routing persists. Most submitted platforms (CorPipe structures, Stanza, AUKBC-MULCRF, LLM-PortNLP) prefer consolidating weights cross-lingually to formulate powerful general-domain models seamlessly reacting to implicit syntax. For instance, DAggerCoref trains a single joint model across all 27 datasets; however, the authors noted this monolithic approach struggled with ancient languages (e.g., Ancient Hebrew, Old Church Slavonic) due to the limited exposure of the underlying XLM-R encoder to classical corpora. Breaking from this purely global strategy, LLM-LatticeNLP represents a noticeable hybrid progression, establishing sequential staging wherein a broad format-aware foundational epoch dictates general coreference metrics, structurally succeeded by dedicated individual language/dataset-specific adapters continuously trained over separate discrete epochs. LLM-UWB similarly compartmentalizes domain behaviors based heavily around whether document bounds enforce the requirement for a separate computational scrolling-context subset fine-tune. Only CorPipeEnsemble relied strictly on parallel inference scaling dynamics, processing outcomes simultaneously across seven identically structured \texttt{umT5-XXL} instances.

\section{Results and Comparison}
\label{sec:results}

\begin{table*}\centering
\begin{tabular}{@{}l r r@{~~}l r@{~~}l r@{~~}l @{}}\toprule
                    & \MC{5}{excluding singletons} & \MC{2}{with singletons}\\\cmidrule(lr){2-6}\cmidrule(l){7-8}
\bf system          & \bf head-match & \MC{2}{\bf partial-match} & \MC{2}{\bf exact-match} & \MC{2}{\bf head-match}\\\midrule
LLM-LatticeNLP      & \bf 74.32 & \bf 74.32 & (-0.00) &     41.83 & (-32.48) & \bf 76.09 & (+1.77)\\
LLM-UWB             &     73.83 &     73.83 & (+0.00) &     40.76 & (-33.08) &     75.59 & (+1.75)\\
LLM-PortNLP         &     68.69 &     67.51 & (-1.18) & \bf 65.30 &  (-3.39) &     70.98 & (+2.29)\\
LLM-Landcore        &     46.19 &     44.79 & (-1.40) &     40.47 &  (-5.73) &     47.78 & (+1.59)\\\midrule
CorPipeEnsemble     & \bf 77.11 & \bf 76.30 & (-0.81) & \bf 74.07 &  (-3.04) & \bf 79.11 & (+2.00)\\
CorPipeXXL          &     76.18 &     75.32 & (-0.86) &     72.90 &  (-3.28) &     78.15 & (+1.97)\\
CorPipeLarge        &     72.32 &     71.16 & (-1.16) &     68.79 &  (-3.53) &     74.52 & (+2.20)\\
DAggerCoref         &     67.56 &     67.56 & (+0.00) &     37.63 & (-29.94) &     58.75 & (-8.81)\\
Stanza              &     67.00 &     65.87 & (-1.13) &     63.32 & (-3.68) &     68.53 & (+1.54)\\
\baselinegz         &     55.39 &     55.06 & (-0.33) &     53.91 & (-1.49) &     48.28 & (-7.11)\\
\baseline           &     54.54 &     54.16 & (-0.38) &     52.96 & (-1.58) &     47.45 & (-7.09)\\
AUKBC-MULCRF        &     35.24 &     35.21 & (-0.03) &     20.65 & (-14.59) &     33.10 & (-2.14)\\\midrule
\sys{Winner-2023}   &     74.90 &     73.33 & (-1.57) &     71.46 & (-3.44) &     76.82 & (+1.91)\\
\sys{Winner-2024}   &     73.90 &     72.19 & (-1.71) &     69.86 & (-4.04) &     75.65 & (+1.75)\\
\sys{Winner-2025}   &     75.84 &     74.90 & (-0.94) &     72.76 & (-3.08) &     78.33 & (+2.49)\\
\sys{Baseline-2023} &     56.96 &     56.28 & (-0.68) &     54.75 & (-2.21) &     49.32 & (-7.64)\\
\sys{Baseline-2024} &     53.16 &     52.48 & (-0.68) &     51.26 & (-1.90) &     46.45 & (-6.71)\\
\sys{Baseline-2025} &     56.01 &     55.58 & (-0.43) &     54.24 & (-1.77) &     47.88 & (-8.13)\\
\bottomrule\end{tabular}

\caption{\textbf{Main results}: the CoNLL F$_1$ score macro-averaged over all datasets.
The table shows the primary metric (head-match excluding singletons)
 and three alternative metrics:
 partial-match excluding singletons,
 exact-match excluding singletons and
 head-match with singletons.
A difference relative to the primary metric is reported in parenthesis.
The top section shows the LLM track, below is the Unconstrained track.
The best score in each column and each of these two sections is in bold.
The systems are ordered by the primary metric.
Similar notes apply to the following tables.
The last six rows showing the winner and baseline results from CRAC~2023--2025
 are copied from the last year Findings \citep{oursharedtask2025},
 and thus are not directly comparable with the rest of the table
 because both the test and training data have been changed (CorefUD~1.1 vs. 1.2 vs. 1.3 vs. 1.4).
}
\label{tab:main-results}
\end{table*}

\begin{table*}\centering
\resizebox{\textwidth}{!}{
\begin{tabular}{@{}l ccccccc @{}}\toprule
\bf system        & \bf MUC &\bf B$^3$ &\bf CEAF-e &\bf BLANC &\bf LEA &\bf MOR &\bf MD-h\\\midrule
CorPipeEnsemble   &  {\bf 82} / {\bf 84} / {\bf 83}  &       73  / {\bf 78} / {\bf 75}  &  {\bf 74} /      72  / {\bf 73}  &  {\bf 74} / {\bf 78} / {\bf 75}  &  {\bf 71} / {\bf 76} / {\bf 73}  &       81  /      85  / {\bf 82}  &  {\bf 87} / {\bf 88} / {\bf 87} \\
CorPipeXXL        &       82  /      83  /      82   &  {\bf 73} /      76  /      74   &       73  /      72  /      72   &       73  /      76  /      74   &       71  /      73  /      72   &  {\bf 81} /      84  /      82   &       86  /      87  /      87  \\
LLM-LatticeNLP    &       79  /      83  /      81   &       72  /      76  /      74   &       65  / {\bf 73} /      69   &       72  /      76  /      74   &       70  /      74  /      71   &       42  / {\bf 88} /      55   &       81  /      86  /      84  \\
LLM-UWB           &       80  /      82  /      81   &       71  /      73  /      72   &       68  /      71  /      69   &       72  /      74  /      73   &       68  /      71  /      69   &       42  /      84  /      55   &       81  /      83  /      82  \\
CorPipeLarge      &       80  /      80  /      80   &       69  /      71  /      70   &       69  /      67  /      68   &       70  /      71  /      70   &       66  /      68  /      67   &       79  /      80  /      79   &       85  /      85  /      85  \\
LLM-PortNLP       &       74  /      82  /      78   &       64  /      71  /      67   &       58  /      69  /      62   &       64  /      73  /      67   &       61  /      69  /      65   &       68  /      82  /      74   &       74  /      83  /      78  \\
DAggerCoref       &       73  /      80  /      77   &       60  /      71  /      65   &       59  /      65  /      62   &       61  /      71  /      65   &       57  /      68  /      62   &       38  /      83  /      51   &       75  /      82  /      78  \\
Stanza            &       74  /      79  /      76   &       63  /      68  /      65   &       61  /      62  /      61   &       63  /      69  /      65   &       60  /      65  /      62   &       72  /      81  /      75   &       79  /      84  /      81  \\
\baselinegz       &       60  /      75  /      66   &       46  /      61  /      51   &       46  /      56  /      49   &       47  /      62  /      51   &       42  /      56  /      47   &       55  /      86  /      65   &       69  /      87  /      76  \\
\baseline         &       60  /      73  /      65   &       45  /      60  /      50   &       45  /      55  /      49   &       45  /      61  /      50   &       41  /      55  /      46   &       54  /      85  /      65   &       68  /      85  /      74  \\
LLM-Landcore      &       48  /      66  /      54   &       38  /      56  /      43   &       39  /      52  /      42   &       33  /      54  /      37   &       34  /      52  /      39   &       49  /      72  /      56   &       54  /      74  /      61  \\
AUKBC-MULCRF      &       46  /      43  /      44   &       31  /      34  /      31   &       36  /      28  /      31   &       31  /      30  /      28   &       24  /      30  /      26   &       35  /      60  /      43   &       63  /      56  /      59  \\
\bottomrule\end{tabular}
}
\caption{Recall / Precision / F1 for individual \textbf{secondary metrics}.
All scores macro-averaged over all datasets.
Both MOR and MD-h measure mention detection quality, while ignoring the coreference.
The MD-h metric (mention detection score on heads) measures only the mention matching quality,
while MOR considers also the whole span detection.
}
\label{tab:secondary-metrics}
\end{table*}

\paragraph{Main results}

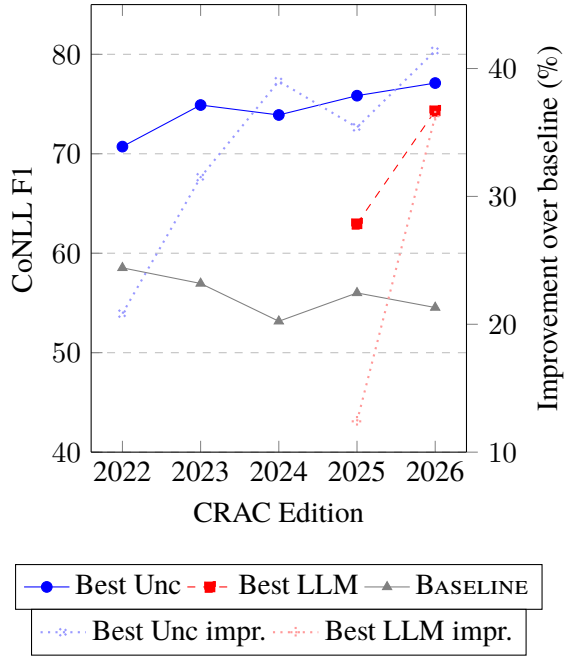
\begin{figure}\centering
\begin{tikzpicture}
  \begin{axis}[
    xlabel={CRAC Edition},
    ylabel={CoNLL F1},
    ymin=40, ymax=85,
    width=0.85\columnwidth, height=7.5cm,
    xtick={1, 2, 3, 4, 5},
    xticklabels={2022, 2023, 2024, 2025, 2026},
    legend style={at={(0.5, -0.25)}, anchor=north},
    legend columns=4,
    ymajorgrids=true,
    grid style=dashed,
    axis y line*=left,
  ]
    \addplot [solid, color=blue, mark=*] coordinates {
      (1, 70.72)
      (2, 74.9)
      (3, 73.9)
      (4, 75.84)
      (5, 77.11)
    };
    \addlegendentry{Best Unc}
    \addplot [dashed, color=red, mark=square*] coordinates {
      (4, 62.96)
      (5, 74.32)
    };
    \addlegendentry{Best LLM}
    \addplot [solid, color=gray, mark=triangle*] coordinates {
      (1, 58.53)
      (2, 56.96)
      (3, 53.16)
      (4, 56.01)
      (5, 54.54)
    };
    \addlegendentry{\baseline{}}
  \end{axis}
  \begin{axis}[
    ylabel={Improvement over baseline (\%)},
    ymin=10, ymax=45,
    width=0.85\columnwidth, height=7.5cm,
    xtick={1, 2, 3, 4, 5},
    xticklabels={2022, 2023, 2024, 2025, 2026},
    legend style={at={(0.5, -0.35)}, anchor=north},
    legend columns=4,
    axis y line*=right,
    axis x line=none,
    ytick align=outside,
  ]
    \addplot [dotted, color=blue!40, mark=x, thick] coordinates {
      (1, 20.83)
      (2, 31.5)
      (3, 39.01)
      (4, 35.4)
      (5, 41.38)
    };
    \addlegendentry{Best Unc impr.}
    \addplot [dotted, color=red!40, mark=+, thick] coordinates {
      (4, 12.41)
      (5, 36.27)
    };
    \addlegendentry{Best LLM impr.}
  \end{axis}
\end{tikzpicture}
\caption{CoNLL F1 scores of the best systems and the baseline across CRAC editions (left axis), and relative improvement over \baseline{} in \% (right axis).}
\label{fig:scores-over-years}
\end{figure}

The main results of the shared task are presented in Table~\ref{tab:main-results}.
Systems LLM-LatticeNLP and CorPipeEnsemble won the LLM and Unconstrained tracks, achieving primary scores of 74.32 and 77.11, respectively.
Generative LLMs thus still could not outperform the best discriminative encoder-based systems, but the gap has narrowed to less than 3 points (compared to 13 points last year), see Figure~\ref{fig:scores-over-years}.

Looking at the LLM track only, one can clearly see the difference made by fine-tuning.
Even though LLM-Landcore uses a frontier-level closed-source model, relying on in-context learning alone was not sufficient to surpass \baseline{} performance.
By contrast, the fine-tuned open-weight LLM approaches remained highly competitive up to the very last moment (see Appendix~\ref{sec:codalab-evol}) and achieved performance comparable to the encoder-based models.

As in previous editions, the CorPipe team and their architecture dominate the Unconstrained track, outperforming the other encoder-based systems by roughly 10 points, including the widely used Stanza system.
AUKBC-MULCRF ranks lower in absolute terms, but offers a notably lightweight alternative to large neural models.
Although the authors do not report the exact number of trainable parameters, the system is likely, together with Stanza (Table~\ref{tab:context-size}), among the most parameter-efficient submissions. This observation broadly aligns with the overall trend that larger models tend to achieve higher scores.

\paragraph{Secondary metrics}
Table~\ref{tab:main-results} also reports CoNLL F$_1$ scores under alternative mention-matching strategies.
While the system ranking under partial match broadly mirrors head match, exact match more clearly distinguishes systems that resolve only mention heads rather than full spans (LLM-LatticeNLP, LLM-UWB, and DAggerCoref), as indicated by the absence of mentions longer than one token in Table~\ref{tab:stats-mentions-nonsingleton} in Appendix~\ref{sec:stats-concat}.
For these systems, exact-match CoNLL scores are about 30 points lower than the corresponding head-match scores.
This aligns with the MOR results in Table~\ref{tab:secondary-metrics}, which fall well below \baseline{} for these systems; we do not observe comparable drops in MD-h.

Including singletons (single-mention entities) in the head-match evaluation in Table~\ref{tab:main-results} increases CoNLL scores for all participating systems except DAggerCoref and AUKBC-MULCRF, suggesting that these systems do not model singletons.
This is confirmed by the absence (or only a marginal number) of singletons in Table~\ref{tab:stats-entities} in Appendix~\ref{sec:stats-concat}.

Table~\ref{tab:secondary-metrics} reports additional coreference measures; aside from minor differences, they yield the same ranking as the primary score
(except for MOR, where LLM-LatticeNLP, LLM-UWB, and DAggerCoref are penalized in recall for predicting only mention heads, as mentioned above).
The interpretation of recall and precision in coreference evaluation is tricky and differs for each metric.
That said, we can notice that the top-scoring systems (both LLM and unconstrained)
maintain relatively small gaps between precision and recall.
The other systems mostly show higher precision than recall,
but the difference is not as striking as it was last year.

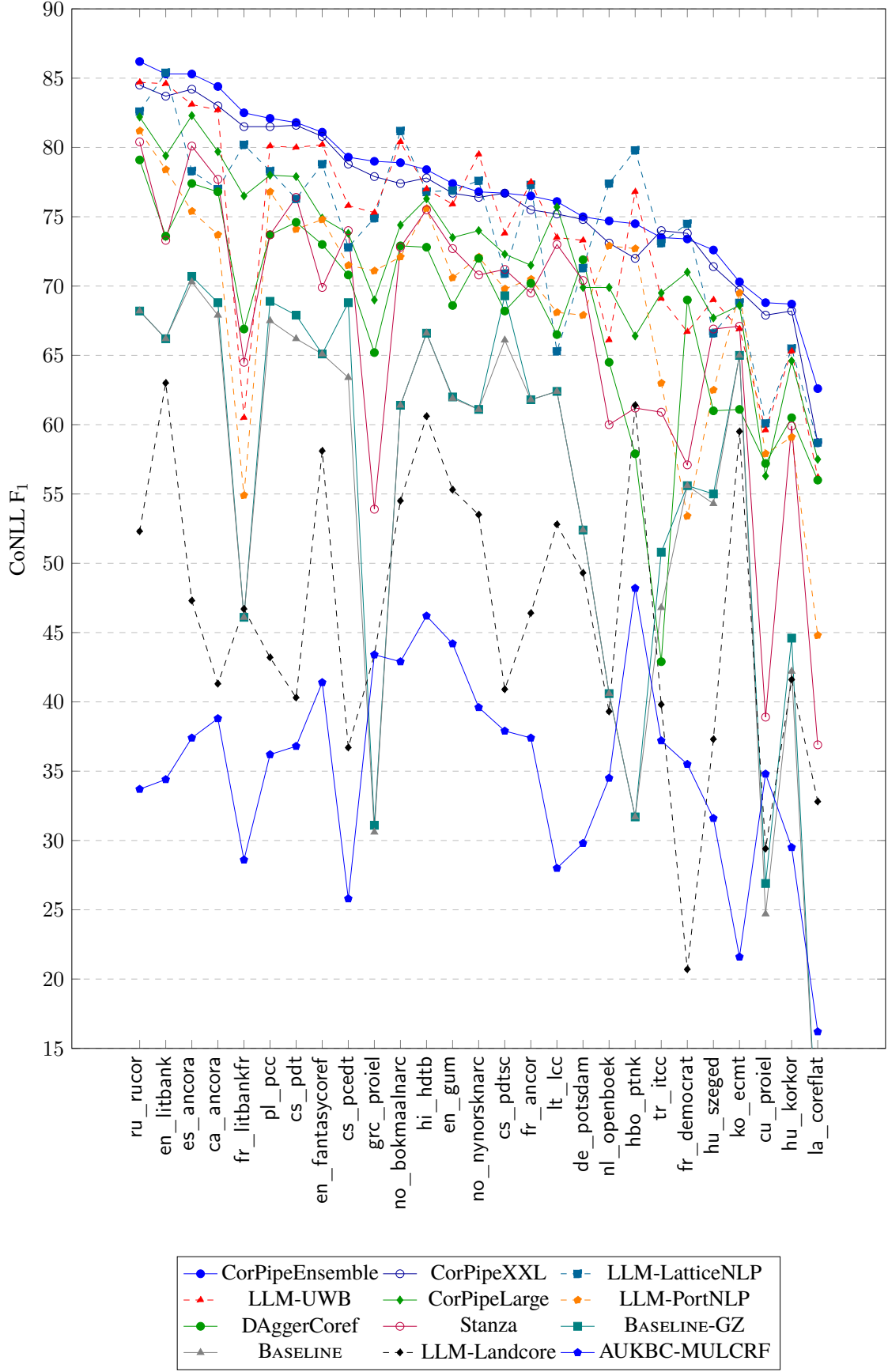
\begin{figure*}\centering
\resizebox{!}{23cm}{
\begin{tikzpicture}
  \begin{axis}[
    ylabel={CoNLL F$_1$},
    ymin=15, ymax=90,
    width=\textwidth, height=20cm,
    xtick={1, 2, 3, 4, 5, 6, 7, 8, 9, 10, 11, 12, 13, 14, 15, 16, 17, 18, 19, 20, 21, 22, 23, 24, 25, 26, 27},
    xticklabels={\rot\rurucor, \rot\enlitbank, \rot\esancora, \rot\caancora, \rot\frlitbankfr, \rot\plpcc, \rot\cspdt, \rot\enfantasycoref, \rot\cspcedt, \rot\grcproiel, \rot\nobokmaalnarc, \rot\hihdtb, \rot\engum, \rot\nonynorsknarc, \rot\cspdtsc, \rot\francor, \rot\ltlcc, \rot\depotsdamcc, \rot\nlopenboek, \rot\hboptnk, \rot\tritcc, \rot\frdemocrat, \rot\huszegedkoref, \rot\koecmt, \rot\cuproiel, \rot\hukorkor, \rot\lacoreflat},
    legend style={at={(0.5, -0.2)}, anchor=north},
    legend columns=3,
    ymajorgrids=true,
    grid style=dashed
  ]
    \addplot [solid, color=blue, mark=*] coordinates {
      (1, 86.2)
      (2, 85.3)
      (3, 85.3)
      (4, 84.4)
      (5, 82.5)
      (6, 82.1)
      (7, 81.8)
      (8, 81.1)
      (9, 79.3)
      (10, 79.0)
      (11, 78.9)
      (12, 78.4)
      (13, 77.4)
      (14, 76.8)
      (15, 76.7)
      (16, 76.5)
      (17, 76.1)
      (18, 75.0)
      (19, 74.7)
      (20, 74.5)
      (21, 73.5)
      (22, 73.4)
      (23, 72.6)
      (24, 70.3)
      (25, 68.8)
      (26, 68.7)
      (27, 62.6)
    };
    \addlegendentry{CorPipeEnsemble}
    \addplot [solid, color=blue!60!black, mark=o] coordinates {
      (1, 84.5)
      (2, 83.7)
      (3, 84.2)
      (4, 83.0)
      (5, 81.5)
      (6, 81.5)
      (7, 81.6)
      (8, 80.8)
      (9, 78.8)
      (10, 77.9)
      (11, 77.4)
      (12, 77.8)
      (13, 76.7)
      (14, 76.4)
      (15, 76.7)
      (16, 75.5)
      (17, 75.2)
      (18, 74.8)
      (19, 73.1)
      (20, 72.0)
      (21, 74.0)
      (22, 73.8)
      (23, 71.4)
      (24, 69.7)
      (25, 67.9)
      (26, 68.2)
      (27, 58.7)
    };
    \addlegendentry{CorPipeXXL}
    \addplot [dashed, color=blue!60!green, mark=square*] coordinates {
      (1, 82.6)
      (2, 85.4)
      (3, 78.3)
      (4, 77.0)
      (5, 80.2)
      (6, 78.3)
      (7, 76.3)
      (8, 78.8)
      (9, 72.8)
      (10, 74.9)
      (11, 81.2)
      (12, 76.8)
      (13, 76.9)
      (14, 77.6)
      (15, 70.9)
      (16, 77.3)
      (17, 65.3)
      (18, 71.3)
      (19, 77.4)
      (20, 79.8)
      (21, 73.1)
      (22, 74.5)
      (23, 66.6)
      (24, 68.8)
      (25, 60.1)
      (26, 65.5)
      (27, 58.7)
    };
    \addlegendentry{LLM-LatticeNLP}
    \addplot [dashed, color=red, mark=triangle*] coordinates {
      (1, 84.7)
      (2, 84.6)
      (3, 83.1)
      (4, 82.7)
      (5, 60.5)
      (6, 80.1)
      (7, 80.0)
      (8, 80.2)
      (9, 75.8)
      (10, 75.3)
      (11, 80.4)
      (12, 77.0)
      (13, 75.9)
      (14, 79.5)
      (15, 73.8)
      (16, 77.5)
      (17, 73.5)
      (18, 73.3)
      (19, 66.1)
      (20, 76.8)
      (21, 69.1)
      (22, 66.7)
      (23, 69.0)
      (24, 66.9)
      (25, 59.6)
      (26, 65.3)
      (27, 56.2)
    };
    \addlegendentry{LLM-UWB}
    \addplot [solid, color=green!60!black, mark=diamond*] coordinates {
      (1, 82.2)
      (2, 79.4)
      (3, 82.3)
      (4, 79.7)
      (5, 76.5)
      (6, 78.0)
      (7, 77.9)
      (8, 74.9)
      (9, 73.8)
      (10, 69.0)
      (11, 74.4)
      (12, 76.3)
      (13, 73.5)
      (14, 74.0)
      (15, 72.3)
      (16, 71.5)
      (17, 75.7)
      (18, 69.9)
      (19, 69.9)
      (20, 66.4)
      (21, 69.5)
      (22, 71.0)
      (23, 67.7)
      (24, 68.6)
      (25, 56.3)
      (26, 64.6)
      (27, 57.5)
    };
    \addlegendentry{CorPipeLarge}
    \addplot [dashed, color=orange, mark=pentagon*] coordinates {
      (1, 81.2)
      (2, 78.4)
      (3, 75.4)
      (4, 73.7)
      (5, 54.9)
      (6, 76.8)
      (7, 74.1)
      (8, 74.8)
      (9, 71.5)
      (10, 71.1)
      (11, 72.1)
      (12, 75.6)
      (13, 70.6)
      (14, 72.1)
      (15, 69.8)
      (16, 70.5)
      (17, 68.1)
      (18, 67.9)
      (19, 72.9)
      (20, 72.7)
      (21, 63.0)
      (22, 53.4)
      (23, 62.5)
      (24, 69.5)
      (25, 57.9)
      (26, 59.1)
      (27, 44.8)
    };
    \addlegendentry{LLM-PortNLP}
    \addplot [solid, color=green!60!black, mark=*] coordinates {
      (1, 79.1)
      (2, 73.6)
      (3, 77.4)
      (4, 76.8)
      (5, 66.9)
      (6, 73.7)
      (7, 74.6)
      (8, 73.0)
      (9, 70.8)
      (10, 65.2)
      (11, 72.9)
      (12, 72.8)
      (13, 68.6)
      (14, 72.0)
      (15, 68.2)
      (16, 70.2)
      (17, 66.5)
      (18, 71.9)
      (19, 64.5)
      (20, 57.9)
      (21, 42.9)
      (22, 69.0)
      (23, 61.0)
      (24, 61.1)
      (25, 57.2)
      (26, 60.5)
      (27, 56.0)
    };
    \addlegendentry{DAggerCoref}
    \addplot [solid, color=purple, mark=o] coordinates {
      (1, 80.4)
      (2, 73.3)
      (3, 80.1)
      (4, 77.7)
      (5, 64.5)
      (6, 73.7)
      (7, 76.4)
      (8, 69.9)
      (9, 74.0)
      (10, 53.9)
      (11, 72.8)
      (12, 75.5)
      (13, 72.7)
      (14, 70.8)
      (15, 71.2)
      (16, 69.5)
      (17, 73.0)
      (18, 70.4)
      (19, 60.0)
      (20, 61.2)
      (21, 60.9)
      (22, 57.1)
      (23, 66.9)
      (24, 67.1)
      (25, 38.9)
      (26, 59.9)
      (27, 36.9)
    };
    \addlegendentry{Stanza}
    \addplot [solid, color=teal, mark=square*] coordinates {
      (1, 68.2)
      (2, 66.2)
      (3, 70.7)
      (4, 68.8)
      (5, 46.1)
      (6, 68.9)
      (7, 67.9)
      (8, 65.1)
      (9, 68.8)
      (10, 31.1)
      (11, 61.4)
      (12, 66.6)
      (13, 62.0)
      (14, 61.1)
      (15, 69.3)
      (16, 61.8)
      (17, 62.4)
      (18, 52.4)
      (19, 40.6)
      (20, 31.7)
      (21, 50.8)
      (22, 55.6)
      (23, 55.0)
      (24, 65.0)
      (25, 26.9)
      (26, 44.6)
      (27, 6.8)
    };
    \addlegendentry{\baselinegz}
    \addplot [solid, color=gray, mark=triangle*] coordinates {
      (1, 68.2)
      (2, 66.2)
      (3, 70.3)
      (4, 67.9)
      (5, 46.1)
      (6, 67.5)
      (7, 66.2)
      (8, 65.1)
      (9, 63.4)
      (10, 30.6)
      (11, 61.4)
      (12, 66.6)
      (13, 61.9)
      (14, 61.1)
      (15, 66.1)
      (16, 61.8)
      (17, 62.4)
      (18, 52.4)
      (19, 40.6)
      (20, 31.7)
      (21, 46.8)
      (22, 55.6)
      (23, 54.3)
      (24, 65.0)
      (25, 24.7)
      (26, 42.2)
      (27, 6.8)
    };
    \addlegendentry{\baseline}
    \addplot [dashed, color=black, mark=diamond*] coordinates {
      (1, 52.3)
      (2, 63.0)
      (3, 47.3)
      (4, 41.3)
      (5, 46.7)
      (6, 43.2)
      (7, 40.3)
      (8, 58.1)
      (9, 36.7)
      (10, 43.4)
      (11, 54.5)
      (12, 60.6)
      (13, 55.3)
      (14, 53.5)
      (15, 40.9)
      (16, 46.4)
      (17, 52.8)
      (18, 49.3)
      (19, 39.3)
      (20, 61.4)
      (21, 39.8)
      (22, 20.7)
      (23, 37.3)
      (24, 59.5)
      (25, 29.4)
      (26, 41.6)
      (27, 32.8)
    };
    \addlegendentry{LLM-Landcore}
    \addplot [solid, color=blue, mark=pentagon*] coordinates {
      (1, 33.7)
      (2, 34.4)
      (3, 37.4)
      (4, 38.8)
      (5, 28.6)
      (6, 36.2)
      (7, 36.8)
      (8, 41.4)
      (9, 25.8)
      (10, 43.4)
      (11, 42.9)
      (12, 46.2)
      (13, 44.2)
      (14, 39.6)
      (15, 37.9)
      (16, 37.4)
      (17, 28.0)
      (18, 29.8)
      (19, 34.5)
      (20, 48.2)
      (21, 37.2)
      (22, 35.5)
      (23, 31.6)
      (24, 21.6)
      (25, 34.8)
      (26, 29.5)
      (27, 16.2)
    };
    \addlegendentry{AUKBC-MULCRF}
  \end{axis}
\end{tikzpicture}
}
\caption{Plot with results for individual languages in the primary metric (CoNLL F$_1$).
This plot shows the same information as Table~\ref{tab:all-langs},
but languages are sorted according to the performance of the best system
and LLM-based systems are shown with dashed lines.}
\label{fig:alllangs}
\end{figure*}

\paragraph{Comparison across datasets}
Table~\ref{tab:all-langs} and Figure~\ref{fig:alllangs} both show CoNLL F$_1$ scores of all systems across the datasets.
To make patterns more visible, the datasets in Figure~\ref{fig:alllangs} are ordered from left to right by the decreasing performance of the top system, CorPipeEnsemble.
Last year, the CorPipe team was outperformed on two datasets only:
on \hboptnk{} by LLM-NUST-FewShot (+10.4) and on \enlitbank{} by LLM-UWB (+2.2).
This year, it has been outperformed on 7 datasets, most notably on the following three:
on \hboptnk{} by LLM-LatticeNLP (+5.3) and LLM-UWB (+2.3),
on \nonynorsknarc{} by LLM-UWB (+2.7) and LLM-LatticeNLP (+0.8), and
on \nlopenboek{} by LLM-LatticeNLP (+2.7).
Similarly to last year, we are not sure about the cause of these differences.
Figure~\ref{fig:alllangs} shows that all systems, except for the three CorPipes and LLM-LatticeNLP, struggled with \frlitbankfr{} (relative to CorPipeEnsemble)
, and most of them also struggled with \grcproiel{}.

\paragraph{Performance on zero mentions}

Table~\ref{tab:all-langs-zero} reports system performance on datasets containing zero mentions, evaluated with the anaphor-decomposable score for zero anaphora.
In CorefUD 1.4, we extended this set of datasets by \cspdtsc{}.

Looking at the F1 scores, CorPipe submissions dominate in most languages.
The only exception is \hukorkor{}, where LLM-UWB and LLM-LatticeNLP score about 10 points higher.
We do not observe this pattern on \huszegedkoref{}, where empty nodes are less frequent (relative to the number of words) and were annotated less rigorously.
Annotation guidelines have a strong impact on zero-mention evaluation, as also seen in the Czech datasets, where none of the systems surpass \baselinegz{}.
Although pro-drop in Czech is not, from a linguistic perspective, considered more complex than in Hungarian or Turkish, the Czech annotation guidelines make resolution harder.
LLM-Landcore and AUKBC-MULCRF have notably worse results on zero anaphors.\footnote{
 AUKBC-MULCRF used the empty-nodes baseline starting point,
  so empty nodes are presented in its predictions for all the 11 datasets,
  but for \hukorkor{}, \huszegedkoref{}, and \plpcc{}, there are no zero anaphora predicted.
 LLM-Landcore had to predict the empty nodes from scratch (the empty-nodes baseline is allowed only in the unconstrained track), and in \caancora{} it predicted no empty nodes.
}

Compared to last year, we observe positive progress: this year's winning system performs similarly, or even better (e.g., on \cspcedt{} and \grcproiel{}) than last year's winner.

\paragraph{Comparison over years}
With five iterations of the shared task now completed, we can examine longitudinal trends in multilingual coreference resolution.
Although the data and some task details have changed from year to year, the same baseline approach has been retained throughout: a fixed baseline system retrained each year on the current release.
This provides a consistent reference point for measuring progress by relating each year's top submissions to \baseline{}.

Figure~\ref{fig:scores-over-years} tracks the best unconstrained and best LLM-based submissions over time.
After the dip observed last year (likely driven in part by the removal of the ParCorFull datasets), the top unconstrained system rebounds and now surpasses \baseline{} by more than 41~\%.
The LLM Track also improves markedly: the strongest LLM-based system increases its gain over \baseline{} from 12~\% in the last year to 36~\% this year.
Overall, the trend is clear: despite the benchmark becoming tougher as new datasets are added, state-of-the-art systems keep pushing performance upward year after year.

\begin{table*}\centering
\resizebox{\textwidth}{!}{
\begin{tabular}{@{}l r@{~~~}r@{~~~}r@{~~~}r@{~~~}r@{~~~}r@{~~~}r@{~~~}r@{~~~}r@{~~~}r@{~~~}r@{~~~}r@{~~~}r@{~~~}r@{~~~}r@{~~~}r@{~~~}r@{~~~}r@{~~~}r@{~~~}r@{~~~}r@{~~~}r@{~~~}r@{~~~}r@{~~~}r@{~~~}r@{~~~}r@{}}\toprule
\bf system        & \rot\caancora & \rot\cspcedt & \rot\cspdt & \rot\cspdtsc & \rot\cuproiel & \rot\depotsdamcc & \rot\enfantasycoref & \rot\engum & \rot\enlitbank & \rot\esancora & \rot\francor & \rot\frdemocrat & \rot\frlitbankfr & \rot\grcproiel & \rot\hboptnk & \rot\hihdtb & \rot\hukorkor & \rot\huszegedkoref & \rot\koecmt & \rot\lacoreflat & \rot\ltlcc & \rot\nlopenboek & \rot\nobokmaalnarc & \rot\nonynorsknarc & \rot\plpcc & \rot\rurucor & \rot\tritcc\\\midrule
CorPipeEnsemble   &\bf  84.4 &\bf  79.3 &\bf  81.8 &\bf  76.7 &\bf  68.8 &\bf  75.0 &\bf  81.1 &\bf  77.4 &     85.3 &\bf  85.3 &     76.5 &     73.4 &\bf  82.5 &\bf  79.0 &     74.5 &\bf  78.4 &\bf  68.7 &\bf  72.6 &\bf  70.3 &\bf  62.6 &\bf  76.1 &     74.7 &     78.9 &     76.8 &\bf  82.1 &\bf  86.2 &     73.5\\
CorPipeXXL        &     83.0 &     78.8 &     81.6 &     76.7 &     67.9 &     74.8 &     80.8 &     76.7 &     83.7 &     84.2 &     75.5 &     73.8 &     81.5 &     77.9 &     72.0 &     77.8 &     68.2 &     71.4 &     69.7 &     58.7 &     75.2 &     73.1 &     77.4 &     76.4 &     81.5 &     84.5 &\bf  74.0\\
LLM-LatticeNLP    &     77.0 &     72.8 &     76.3 &     70.9 &     60.1 &     71.3 &     78.8 &     76.9 &\bf  85.4 &     78.3 &     77.3 &\bf  74.5 &     80.2 &     74.9 &\bf  79.8 &     76.8 &     65.5 &     66.6 &     68.8 &     58.7 &     65.3 &\bf  77.4 &\bf  81.2 &     77.6 &     78.3 &     82.6 &     73.1\\
LLM-UWB           &     82.7 &     75.8 &     80.0 &     73.8 &     59.6 &     73.3 &     80.2 &     75.9 &     84.6 &     83.1 &\bf  77.5 &     66.7 &     60.5 &     75.3 &     76.8 &     77.0 &     65.3 &     69.0 &     66.9 &     56.2 &     73.5 &     66.1 &     80.4 &\bf  79.5 &     80.1 &     84.7 &     69.1\\
CorPipeLarge      &     79.7 &     73.8 &     77.9 &     72.3 &     56.3 &     69.9 &     74.9 &     73.5 &     79.4 &     82.3 &     71.5 &     71.0 &     76.5 &     69.0 &     66.4 &     76.3 &     64.6 &     67.7 &     68.6 &     57.5 &     75.7 &     69.9 &     74.4 &     74.0 &     78.0 &     82.2 &     69.5\\
LLM-PortNLP       &     73.7 &     71.5 &     74.1 &     69.8 &     57.9 &     67.9 &     74.8 &     70.6 &     78.4 &     75.4 &     70.5 &     53.4 &     54.9 &     71.1 &     72.7 &     75.6 &     59.1 &     62.5 &     69.5 &     44.8 &     68.1 &     72.9 &     72.1 &     72.1 &     76.8 &     81.2 &     63.0\\
DAggerCoref       &     76.8 &     70.8 &     74.6 &     68.2 &     57.2 &     71.9 &     73.0 &     68.6 &     73.6 &     77.4 &     70.2 &     69.0 &     66.9 &     65.2 &     57.9 &     72.8 &     60.5 &     61.0 &     61.1 &     56.0 &     66.5 &     64.5 &     72.9 &     72.0 &     73.7 &     79.1 &     42.9\\
Stanza            &     77.7 &     74.0 &     76.4 &     71.2 &     38.9 &     70.4 &     69.9 &     72.7 &     73.3 &     80.1 &     69.5 &     57.1 &     64.5 &     53.9 &     61.2 &     75.5 &     59.9 &     66.9 &     67.1 &     36.9 &     73.0 &     60.0 &     72.8 &     70.8 &     73.7 &     80.4 &     60.9\\
\baselinegz       &     68.8 &     68.8 &     67.9 &     69.3 &     26.9 &     52.4 &     65.1 &     62.0 &     66.2 &     70.7 &     61.8 &     55.6 &     46.1 &     31.1 &     31.7 &     66.6 &     44.6 &     55.0 &     65.0 &      6.8 &     62.4 &     40.6 &     61.4 &     61.1 &     68.9 &     68.2 &     50.8\\
\baseline         &     67.9 &     63.4 &     66.2 &     66.1 &     24.7 &     52.4 &     65.1 &     61.9 &     66.2 &     70.3 &     61.8 &     55.6 &     46.1 &     30.6 &     31.7 &     66.6 &     42.2 &     54.3 &     65.0 &      6.8 &     62.4 &     40.6 &     61.4 &     61.1 &     67.5 &     68.2 &     46.8\\
LLM-Landcore      &     41.3 &     36.7 &     40.3 &     40.9 &     29.4 &     49.3 &     58.1 &     55.3 &     63.0 &     47.3 &     46.4 &     20.7 &     46.7 &     43.4 &     61.4 &     60.6 &     41.6 &     37.3 &     59.5 &     32.8 &     52.8 &     39.3 &     54.5 &     53.5 &     43.2 &     52.3 &     39.8\\
AUKBC-MULCRF      &     38.8 &     25.8 &     36.8 &     37.9 &     34.8 &     29.8 &     41.4 &     44.2 &     34.4 &     37.4 &     37.4 &     35.5 &     28.6 &     43.4 &     48.2 &     46.2 &     29.5 &     31.6 &     21.6 &     16.2 &     28.0 &     34.5 &     42.9 &     39.6 &     36.2 &     33.7 &     37.2\\
\bottomrule\end{tabular}
}
\caption{Results for \textbf{individual languages} in the primary metric (CoNLL F$_1$).
}
\label{tab:all-langs}
\end{table*}

\begin{table*}\centering
\resizebox{\textwidth}{!}{
\begin{tabular}{@{}l r@{~~~}r@{~~~}r@{~~~}r@{~~~}r@{~~~}r@{~~~}r@{~~~}r@{~~~}r@{~~~}r@{~~~}r@{}}\toprule
\bf system        & \rot\caancora & \rot\cspcedt & \rot\cspdt & \rot\cspdtsc & \rot\cuproiel & \rot\esancora & \rot\grcproiel & \rot\hukorkor & \rot\huszegedkoref & \rot\plpcc & \rot\tritcc\\\midrule
CorPipeEnsemble   &       90  /      87  /      88   &       69  /      79  /      74   &       82  /      86  /      84   &       86  /      86  /      86   &       79  /      80  /      80   &       94  /      93  /      93   &  {\bf 88} /      93  / {\bf 90}  &       65  /      80  /      72   &  {\bf 86} /      74  / {\bf 79}  &  {\bf 94} /      86  / {\bf 90}  &       88  /      84  /      86  \\
CorPipeXXL        &       89  /      86  /      88   &       69  /      77  /      73   &       82  /      86  /      84   &       86  /      85  /      86   &  {\bf 79} / {\bf 81} / {\bf 80}  &       94  /      92  /      93   &       88  /      91  /      89   &       63  /      81  /      71   &       85  /      73  /      79   &       93  /      85  /      89   &  {\bf 89} /      84  / {\bf 87} \\
LLM-LatticeNLP    &       83  /      83  /      83   &       64  /      78  /      70   &       77  /      87  /      82   &       78  /      83  /      80   &       77  /      78  /      78   &       89  /      89  /      89   &       87  / {\bf 93} /      90   &       74  / {\bf 86} /      80   &       74  /      73  /      74   &       89  /      85  /      87   &       87  /      85  /      86  \\
LLM-UWB           &  {\bf 90} /      87  /      88   &       70  /      75  /      72   &       78  / {\bf 92} /      85   &       84  /      83  /      84   &       72  /      76  /      74   &       92  / {\bf 94} /      93   &       87  /      88  /      87   &  {\bf 77} /      85  / {\bf 81}  &       78  /      77  /      78   &       89  / {\bf 89} /      89   &       87  /      79  /      83  \\
CorPipeLarge      &       89  /      82  /      86   &       56  /      80  /      66   &       81  /      84  /      82   &       82  /      86  /      84   &       71  /      71  /      71   &  {\bf 94} /      93  / {\bf 93}  &       83  /      88  /      86   &       70  /      74  /      72   &       84  /      65  /      73   &       93  /      82  /      87   &       88  /      80  /      84  \\
LLM-PortNLP       &       89  / {\bf 89} / {\bf 89}  &       62  /      77  /      69   &       80  /      85  /      82   &       81  /      82  /      82   &       76  /      79  /      77   &       91  /      93  /      92   &       88  /      91  /      89   &       57  /      86  /      69   &       63  / {\bf 78} /      70   &       88  /      86  /      87   &       81  / {\bf 87} /      84  \\
DAggerCoref       &       84  /      86  /      85   &       48  /      80  /      60   &       73  /      80  /      76   &       69  /      79  /      74   &       70  /      76  /      73   &       88  /      92  /      90   &       78  /      82  /      80   &       49  /      86  /      63   &       45  /      59  /      51   &       87  /      87  /      87   &       41  /      79  /      54  \\
Stanza            &       88  /      82  /      85   &       56  / {\bf 82} /      67   &       81  /      83  /      82   &       82  /      83  /      82   &       59  /      65  /      62   &       92  /      90  /      91   &       74  /      84  /      79   &       60  /      78  /      68   &       81  /      74  /      77   &       93  /      81  /      87   &       78  /      79  /      79  \\
\baselinegz       &       83  /      82  /      83   &  {\bf 76} /      81  / {\bf 78}  &  {\bf 85} /      87  / {\bf 86}  &  {\bf 89} / {\bf 88} / {\bf 89}  &       57  /      69  /      63   &       89  /      89  /      89   &       63  /      69  /      66   &       59  /      75  /      66   &       51  /      57  /      54   &       90  /      88  /      89   &       76  /      81  /      79  \\
\baseline         &       81  /      74  /      77   &       50  /      74  /      60   &       75  /      78  /      77   &       79  /      83  /      81   &       49  /      58  /      53   &       87  /      86  /      86   &       61  /      69  /      65   &       48  /      60  /      53   &       50  /      53  /      52   &       89  /      79  /      83   &       72  /      71  /      71  \\
LLM-Landcore      &        0  /       0  /       0   &       10  /      39  /      16   &       12  /      46  /      19   &       42  /      44  /      43   &        6  /      10  /       8   &        3  /      39  /       5   &       29  /      61  /      40   &       24  /      51  /      33   &        2  /       9  /       3   &        4  /      40  /       7   &       41  /      55  /      47  \\
AUKBC-MULCRF      &       47  /      50  /      49   &       26  /      37  /      30   &       50  /      54  /      52   &       71  /      65  /      68   &       62  /      61  /      62   &       53  /      58  /      56   &       71  /      72  /      72   &        0  /       0  /       0   &        0  /       0  /       0   &        0  /       0  /       0   &       74  /      65  /      70  \\
\midrule
\sys{Winner-2025} &       91  /      87  /      89   &       61  /      79  /      69   &       82  /      86  /      84   &  --                              &       77  /      80  /      79   &       93  /      92  /      92   &       87  /      87  /      87   &       65  /      81  /      72   &       85  /      73 /       78   &       93  /      84  /      89   &       84  /      83  /      84  \\
\sys{Baseline-2025} &     79  /      75  /      77   &        9  /      93  /      17   &       34  /      82  /      48   &  --                              &       52  /      62  /      57   &       88  /      87  /      87   &       62  /      67  /      64   &       56  /      63  /      59   &       54  /      57  /      55   &       86  /      78  /      82   &       71  /      73  /      72  \\
\bottomrule\end{tabular}
}
\caption{Recall / Precision / F1 for anaphor-decomposable score of coreference resolution on \textbf{zero anaphors} across individual languages.
Only datasets containing anaphoric zeros are listed (\engum{} excluded as all zeros in its test set are non-anaphoric).
Note that these scores are directly comparable to neither the CoNLL score nor the supplementary scores calculated with respect to whole entities in Table~\ref{tab:secondary-metrics}.
}
\label{tab:all-langs-zero}
\end{table*}

\paragraph{Further analysis}
As in previous years, we provide several additional tables in the
appendices to shed more light on the differences between the submitted
systems. Discussions of these tables are provided in their captions.

Tables~\ref{tab:upos-entity}--\ref{tab:upos-mention} in Appendix~\ref{sec:stats-upos} show results factorized
according to the different universal part-of-speech tags (UPOS) in the
mention heads. 

Tables \ref{tab:stats-entities}--\ref{tab:stats-details} in Appendix~\ref{sec:stats-concat} show various
statistics on the entities and mentions in a concatenation of all the test
sets. Note that such statistics are mostly influenced by larger datasets.

\paragraph{Performance on long-range entities}
Appendices \ref{sec:stats-litbankfr} and \ref{sec:stats-openboek} show the same type of tables as Appendix~\ref{sec:stats-concat},
but for the top two long-entity datasets (with the highest p95 range according to Table~\ref{tab:sizes}),
French-LitBankFr and Dutch-OpenBoek, respectively.

In French-LitBankFr, we can see that LLM-UWB, LLM-PortNLP, and Stanza
have p95 ranges (9370, 6729, and 6336, respectively) that are higher than the gold data (4435),
while LLM-Landcore's p95 range is the closest (3575).
All other systems generate entities that are much shorter than those in the gold data.

In Dutch-OpenBoek, LLM-Landcore is again the closest;
Stanza generates longer entities than the gold data;
and all other systems generate much shorter entities.

To better understand how systems handle long-range entities, we conduct a document-level analysis.
We split the official mini-test set into individual documents, compute the primary score for each document, and examine performance as the document p95 range increases.
To reduce noise, we report a sliding-window average of CoNLL F1.
Because short-range documents contain fewer entities, a fixed number of documents per window would yield unstable estimates.
We therefore define windows by a fixed maximum number of tokens.
This procedure is only an approximation: documents with a high p95 range still contain many mid- and short-range entities, which can influence the aggregate score.\footnote{
 We have tried to design a metric that would focus on ``long'' entities only
  (or, in general, on entities of a given range), but we were not successful
  -- e.g., if a system-predicted entity is slightly shorter than the threshold,
  it would be counted as completely missing, which would be too harsh and lead to non-intuitive (unfair) evaluation results.
}
Finally, entity range is not the only driver of document-level performance; language, genre, and annotation guidelines also matter.

Figure~\ref{fig:scores-long-range} presents results for documents whose p95 range exceeds 100 tokens, computed with a 50k-token sliding window.
Systems vary markedly in their ability to handle long-range entities.
Among the best-performing systems, CorPipe and LLM-LatticeNLP remain robust as the range increases, whereas the other systems deteriorate noticeably.
The contrast between LLM-UWB and LLM-LatticeNLP is particularly informative: despite similar overall performance (Table~\ref{tab:main-results}) and a largely shared setup (Table~\ref{tab:context-size}), their behavior diverges across entity ranges.
LLM-UWB performs slightly better on mid-range documents, and the two systems are comparable for p95 ranges between 1k and 2k tokens.
For the longest-range documents, however, LLM-LatticeNLP maintains performance and achieves the best scores overall, while LLM-UWB drops by roughly 15 F1 points.\footnote{
As an alternative to entity range, we have also sorted the documents by the maximum distance between adjacent mentions (in any entity in a given document).
We omit the resulting graph because it is very similar to Figure~\ref{fig:scores-long-range}
-- the CoNLL F1 scores for each document are the same, and the ordering of documents is similar because p95 entity range correlates with the maximum distance between adjacent mentions.
}

\input{scores-long-range}
\section{Conclusions}
\label{sec:conclusions}

The fifth edition of the Shared Task on Multilingual Coreference Resolution successfully expanded the linguistic and structural scope of the CorefUD collection, specifically addressing the challenges posed by long-range entities in literary and spoken datasets. The results demonstrate a significant narrowing of the performance gap between traditional discriminative systems and generative Large Language Models (LLMs). While encoder-based architectures like CorPipe continue to lead in the Unconstrained track, reaching a primary CoNLL $F_1$ score of 77.11, fine-tuned LLMs have emerged as highly competitive alternatives, trailing by less than 3 points—a substantial improvement from the 13-point margin observed in previous years.

Furthermore, the introduction of specialized LLM-based and Unconstrained tracks, along with simplified data formats (plaintext and JSON), facilitated diverse methodological approaches, ranging from few-shot in-context learning to complex hybrid staging. The focus on long-range entities (frequent in the newly added datasets French-LitBankFr, Dutch-OpenBoek, and English-FantasyCoref) highlighted varying system capabilities in this aspect. Overall, these findings underscore the continued evolution of multilingual coreference resolution and the rising efficacy of LLMs in handling complex, long-range linguistic structures.

\section*{Acknowledgements}

\forcamera{update}
This work has been supported
by Charles University Research Centre program No. 24/SSH/009,
Ministry of Education, Youth, and Sports of the Czech Republic, Project No.
LM2023062 LINDAT/CLARIAH-CZ and
CZ.02.01.01/00/23\_020/0008518,
and by the project R\&D of Technologies for Advanced Digitalization in the Pilsen Metropolitan Area (DigiTech) No. CZ.02.01.01/00/23\_021/0008436.
We thank all the participants of the shared task for participating and for
providing brief descriptions of their systems.
We also thank anonymous reviewers for their useful remarks.

\newpage %

\bibliography{anthology,custom}
\bibliographystyle{acl_natbib}

\appendix
\clearpage
\onecolumn

\section{CorefUD 1.4 Details}
\label{sec:data-references}

\begin{table}[H]
  \resizebox{\columnwidth}{!}{
  \begin{tabular}{llll}
    Ancient Greek  & PROIEL      & \grcproiel{}      & \cite{Haug2008CreatingAP} \\
    Ancient Hebrew  & PTNK       & \hboptnk{}        & \cite{swanson-etal-2024-towards} \\
    Catalan    & AnCora          & \caancora{}       & \cite{ancora,ancora-co} \\
    Czech      & PCEDT           & \cspcedt{}        & \cite{PCEDT2016coreference} \\
    Czech      & PDT             & \cspdt{}          & \cite{udpdtc} \\
    Czech      & PDTSC           & \cspdtsc{}        & \cite{pdtsc} \\
    Dutch      & OpenBoek        & \nlopenboek{}     & \cite{Cranenburgh_vanNoord_2022} \\
    English    & FantasyCoref    & \enfantasycoref{} & \cite{han-etal-2021-fantasycoref} \\
    English    & GUM             & \engum{}          & \cite{GUMdocumentation} \\
    English    & LitBank         & \enlitbank{}      & \cite{Bamman20Litbank} \\
    English    & ParCorFull      & \enparcorfull{}   & \cite{ParCorFullScheme} \\
    French     & ANCOR           & \francor{}        & \cite{muzerelle-etal-2014-ancor} \\
    French     & Democrat        & \frdemocrat{}     & \cite{democrat} \\
    French     & LitBankFr       & \frlitbankfr      & \cite{MelanieBecquet2024BookNLPfr} \\
    German     & ParCorFull      & \deparcorfull{}   & \cite{ParCorFullScheme} \\
    German     & PotsdamCC       & \depotsdamcc{}    & \cite{potsdamCC-2020} \\
    Hindi      & HDTB            & \hihdtb{}         & \cite{mujadia-etal-2016-coreference} \\
    Hungarian  & KorKor          & \hukorkor{}       & \cite{korkor_coling} \\
    Hungarian  & SzegedKoref     & \huszegedkoref{}  & \cite{szegedkoref2018} \\
    Korean     & ECMT            & \koecmt{}         & \cite{nam-etal-2020-effective} \\
    Latin      & CorefLat        & \lacoreflat{}     & \cite{delfino2024building} \\
    Lithuanian & LCC             & \ltlcc{}          & \cite{LithuanianScheme} \\
    Norwegian  & Bokmål NARC     & \nobokmaalnarc{}  & \cite{maehlum2022narc} \\
    Norwegian  & Nynorsk NARC    & \nonynorsknarc{}  & \cite{maehlum2022narc} \\
    Old Church Slavonic & PROIEL & \cuproiel         & \cite{Haug2008CreatingAP} \\
    Polish     & PCC             & \plpcc{}          & \cite{PCC2013,bookOgrodniczuk} \\
    Russian    & RuCor           & \rurucor{}        & \cite{RuCorDialog} \\
    Spanish    & AnCora          & \esancora{}       & \cite{ancora,ancora-co} \\
    Turkish    & ITCC            & \tritcc           & \cite{pamay2018coref} \\
  \end{tabular}
  }
  \caption{Datasets in CorefUD 1.4.}
  \label{tab:corefud}
\end{table}

\section{CoNLL results by head UPOS}
\label{sec:stats-upos}

\begin{table}[H]\centering
\begin{tabular}{@{}l r@{~~~}r@{~~~}r@{~~~}r@{~~~}r@{~~~}r@{~~~}r@{~~~}r@{}}\toprule
\bf system         & \bf NOUN  & \bf PRON  & \bf PROPN & \bf DET   & \bf ADJ   & \bf VERB  & \bf ADV   & \bf NUM  \\\midrule
CorPipeEnsemble    & \bf 73.67 & \bf 78.12 & \bf 79.18 & \bf 56.51 & \bf 44.29 & \bf 38.34 &     30.19 & \bf 46.49 \\
CorPipeXXL         &     72.34 &     76.76 &     77.64 &     53.52 &     43.39 &     37.86 & \bf 32.48 &     43.54 \\
LLM-LatticeNLP     &     71.18 &     71.89 &     76.64 &     52.90 &     37.92 &     32.30 &     24.55 &     38.19 \\
LLM-UWB            &     70.80 &     70.40 &     75.77 &     50.46 &     38.01 &     32.51 &     25.67 &     37.21 \\
CorPipeLarge       &     68.29 &     73.17 &     73.35 &     49.15 &     37.63 &     30.57 &     27.54 &     36.95 \\
LLM-PortNLP        &     65.10 &     66.28 &     70.69 &     51.11 &     40.86 &     28.26 &     29.80 &     36.99 \\
DAggerCoref        &     64.28 &     67.68 &     67.11 &     43.41 &     30.55 &     28.37 &     23.32 &     30.23 \\
Stanza             &     62.81 &     67.42 &     67.96 &     43.56 &     32.35 &     25.02 &     23.05 &     31.66 \\
\baselinegz        &     48.35 &     55.61 &     53.77 &     34.58 &     22.45 &     15.02 &     19.80 &     21.17 \\
\baseline          &     47.73 &     54.36 &     53.30 &     33.33 &     22.29 &     14.11 &     20.47 &     21.32 \\
LLM-Landcore       &     41.68 &     41.36 &     50.70 &     24.12 &     19.24 &      6.41 &      8.38 &     11.55 \\
AUKBC-MULCRF       &     25.08 &     22.91 &     29.62 &      9.43 &      3.74 &      4.61 &      4.39 &      5.26 \\
\bottomrule\end{tabular}
\caption[CoNLL by UPOS entities]{CoNLL F$_1$ score (head-match) evaluated only on \textbf{entities} with heads of a given
UPOS. In both the gold and prediction files we deleted some entities before
running the evaluation. We kept only entities with at least one mention with
a given head UPOS (universal part of speech tag). For the purpose of this
analysis,
 if the head node had deprel=flat children,
 their UPOS tags were considered as well,
 so for example in ``Mr./NOUN Brown/PROPN''
 both NOUN and PROPN were taken as head UPOS,
 so the entity with this mention will be reported in both columns NOUN and PROPN.
Otherwise, the CoNLL F$_1$ scores are the same as in the primary metric,
 i.e.\ an unweighted average over all datasets, head-match, without singletons.
Note that when distinguishing entities into events and nominal entities,
 the VERB column can be considered as an approximation of the performance on events.

\hskip5mm The results for each UPOS mostly follow the overall system ordering with a few exceptions,
e.g., CorPipeLarge being the third system in PRON but fifth overall,
or LLM-PortNLP being the third best system in ADJ but sixth best overall.
The easiest to predict are entities with PROPN or PRON,
while the most difficult are entities with ADV or VERB
(which are however quite rare both in the gold and predicted data, cf. Table~\ref{tab:stats-details}).
}
\label{tab:upos-entity}
\end{table}

\begin{table}[H]\centering
\begin{tabular}{@{}l r@{~~~}r@{~~~}r@{~~~}r@{~~~}r@{~~~}r@{~~~}r@{~~~}r@{}}\toprule
\bf system         & \bf NOUN  & \bf PRON  & \bf PROPN & \bf DET   & \bf ADJ   & \bf VERB  & \bf ADV   & \bf NUM  \\\midrule
CorPipeEnsemble    & \bf 64.73 & \bf 70.06 & \bf 64.38 & \bf 55.63 & \bf 51.74 & \bf 51.32 & \bf 52.28 & \bf 51.82 \\
CorPipeXXL         &     63.80 &     69.11 &     63.84 &     54.32 &     50.30 &     50.28 &     51.03 &     50.62 \\
LLM-LatticeNLP     &     61.93 &     63.71 &     62.34 &     53.98 &     49.74 &     49.07 &     49.92 &     49.66 \\
LLM-UWB            &     62.22 &     62.17 &     61.39 &     52.20 &     48.86 &     48.34 &     48.92 &     48.69 \\
CorPipeLarge       &     59.07 &     64.50 &     59.11 &     49.40 &     45.25 &     44.99 &     45.44 &     45.16 \\
LLM-PortNLP        &     55.57 &     57.93 &     57.01 &     48.35 &     44.41 &     43.84 &     44.85 &     44.33 \\
DAggerCoref        &     54.26 &     58.81 &     53.13 &     43.34 &     39.85 &     39.20 &     39.95 &     39.27 \\
Stanza             &     53.31 &     59.14 &     54.70 &     44.42 &     40.62 &     40.17 &     40.97 &     40.56 \\
\baselinegz        &     39.73 &     47.55 &     41.77 &     32.11 &     29.43 &     28.60 &     29.50 &     28.74 \\
\baseline          &     39.17 &     46.41 &     41.09 &     31.24 &     28.38 &     27.64 &     28.59 &     27.73 \\
LLM-Landcore       &     37.46 &     34.88 &     39.47 &     25.77 &     24.87 &     23.76 &     24.61 &     24.35 \\
AUKBC-MULCRF       &     29.90 &     27.87 &     29.28 &     13.29 &     13.49 &     13.64 &     13.94 &     13.92 \\
\bottomrule\end{tabular}
\caption[]{CoNLL F$_1$ score (head-match) evaluated only on \textbf{mentions} with heads of a given UPOS.
In both the gold and prediction files we deleted some mentions before running the evaluation.
We kept only mentions with a given head UPOS (again considering also deprel=flat children).
These results may be a bit
misleading because e.g.\ the PRON column does not consider all pronominal
coreference, but only pronoun-to-pronoun coreference. An entity with one
pronoun and one noun mention is excluded from this table (because it becomes
a singleton after deleting noun or pronoun mentions and singletons are
excluded from the evaluation in this table).

\hskip5mm The results again mostly follow the overall ranking.
The easiest to predict are PRON-to-PRON coreferences.
}
\label{tab:upos-mention}
\end{table}

\section{Statistics of the submitted systems on concatenation of all test sets}
\label{sec:stats-concat}
The systems are sorted alphabetically in the tables in this section.

\begin{table}[H]\centering
\begin{tabular}{@{}l rrrrr rrrrr@{}}\toprule
                    & \MC{5}{entities}                          & \MC{5}{distribution of lengths}     \\\cmidrule(lr){2-6}\cmidrule(l){7-11}
system              &   total & per 1k & \MC{2}{length} & range &    1 &     2 &     3 &     4 &   5+ \\\cmidrule(lr){4-5}
                    &   count &  words &    max &  avg. & p95   & [\%] &  [\%] &  [\%] &  [\%] & [\%] \\\midrule
gold                &  44,756 &    92 &   934 &  2.3 &   344 & 64.5 & 18.6 &  6.3 &  3.0 &  7.6 \\    
AUKBC-MULCRF        &  20,039 &    41 &   714 &  4.3 &   700 &  0.0 & 54.0 & 18.6 &  9.2 & 18.2 \\    
\baseline           &  13,469 &    28 &   411 &  4.8 &   787 &  0.0 & 54.7 & 17.1 &  8.2 & 20.0 \\    
\baselinegz         &  13,493 &    28 &   411 &  4.8 &   782 &  0.0 & 54.5 & 17.2 &  8.2 & 20.1 \\    
CorPipeEnsemble     &  46,670 &    96 &   920 &  2.3 &   332 & 64.4 & 18.4 &  6.5 &  3.3 &  7.4 \\    
CorPipeLarge        &  46,831 &    96 &   919 &  2.3 &   346 & 64.7 & 17.9 &  6.6 &  3.2 &  7.6 \\    
CorPipeXXL          &  46,272 &    95 &   929 &  2.3 &   341 & 64.5 & 18.3 &  6.4 &  3.3 &  7.5 \\    
DAggerCoref         &  14,896 &    31 &   776 &  4.8 &   414 &  2.1 & 48.8 & 18.1 &  9.0 & 22.0 \\    
LLM-Landcore        &  36,075 &    74 &   831 &  2.2 &   387 & 67.0 & 16.7 &  6.0 &  3.1 &  7.2 \\    
LLM-LatticeNLP      &  42,045 &    86 &   974 &  2.4 &   317 & 66.3 & 16.7 &  6.1 &  3.1 &  7.8 \\    
LLM-PortNLP         &  44,689 &    92 &   914 &  2.3 &   283 & 68.9 & 15.7 &  5.5 &  2.8 &  7.2 \\    
LLM-UWB             &  44,355 &    91 &   819 &  2.4 &   394 & 64.8 & 17.6 &  6.4 &  3.3 &  7.9 \\    
Stanza              &  38,825 &    80 &   924 &  2.5 &   313 & 59.4 & 21.7 &  7.2 &  3.4 &  8.3 \\    
\bottomrule\end{tabular}
\caption[]{Statistics on coreference entities.
The five column under the ``entities'' header are described in Table~\ref{tab:sizes}.
Here we add also distribution of entity lengths (singletons have length = 1).

\hskip5mm Note that the gold data contain 15,888 non-singleton entities,
 so the two baselines (that do not predict any singletons) undergenerate
 (i.e.\ predict less non-singleton entities than in the gold data),
 but AUKBC-MULCRF overgenerates (because 20,039$>$15,888).
These three systems and DAggerCoref also predict on average longer entities
 (i.e.\ with more mentions) than in the gold data.
The remaining systems have the statistics similar to the gold data,
 (although the CorPipe* systems slightly overgenerate,
 while LLM-Landcore and Stanza undergenerate).
}
\label{tab:stats-entities}
\end{table}

\begin{table}[H]\centering
\begin{tabular}{@{}l rrrr rrrrrr@{}}\toprule
                    & \MC{4}{non-singleton mentions}     & \MC{6}{distribution of lengths}              \\\cmidrule(lr){2-5}\cmidrule(l){6-11}
system              &   total & per 1k & \MC{2}{length} &     0 &     1 &     2 &     3 &     4 &   5+ \\\cmidrule(lr){4-5}
                    &   count &  words &    max &  avg. &  [\%] &  [\%] &  [\%] &  [\%] &  [\%] & [\%] \\\midrule
gold                &  76,034 &   156 &   100 &  2.2 &  9.5 & 52.5 & 19.4 &  6.5 &  2.9 &  9.2 \\    
AUKBC-MULCRF        &  85,220 &   175 &     6 &  1.0 &  7.1 & 89.6 &  2.8 &  0.4 &  0.1 &  0.0 \\    
\baseline           &  64,442 &   132 &    27 &  1.7 & 10.7 & 56.9 & 18.5 &  5.7 &  2.3 &  5.9 \\    
\baselinegz         &  64,583 &   132 &    27 &  1.7 & 10.9 & 56.8 & 18.4 &  5.7 &  2.3 &  6.0 \\    
CorPipeEnsemble     &  76,559 &   157 &   100 &  2.2 &  9.5 & 53.5 & 19.2 &  6.4 &  2.8 &  8.6 \\    
CorPipeLarge        &  77,022 &   158 &   119 &  2.2 &  9.4 & 53.6 & 19.3 &  6.3 &  2.7 &  8.6 \\    
CorPipeXXL          &  76,778 &   157 &   100 &  2.2 &  9.5 & 53.4 & 19.2 &  6.4 &  2.8 &  8.7 \\    
DAggerCoref         &  70,514 &   145 &     1 &  0.9 &  8.3 & 91.7 &  0.0 &  0.0 &  0.0 &  0.0 \\    
LLM-Landcore        &  55,494 &   114 &    62 &  1.9 &  6.6 & 56.5 & 20.2 &  6.8 &  2.8 &  7.0 \\    
LLM-LatticeNLP      &  72,236 &   148 &     1 &  0.9 &  9.6 & 90.4 &  0.0 &  0.0 &  0.0 &  0.0 \\    
LLM-PortNLP         &  70,571 &   145 &   100 &  2.1 &  9.6 & 55.0 & 19.1 &  5.9 &  2.6 &  7.9 \\    
LLM-UWB             &  75,958 &   156 &     1 &  0.9 &  9.5 & 90.5 &  0.0 &  0.0 &  0.0 &  0.0 \\    
Stanza              &  72,410 &   148 &   104 &  2.1 &  9.6 & 54.6 & 18.8 &  6.1 &  2.6 &  8.3 \\    
\bottomrule\end{tabular}
\caption[]{Statistics on non-singleton mentions.
The total number of mentions and the average number of
mentions per 1000 words of running text. The maximum and average mention length, i.e., the number of nonempty nodes (words) in the mention. Distribution of mention lengths (zeros have length = 0).

\hskip5mm Only the CorPipe* systems and LLM-UWB generate a similar number of non-singleton mentions as in the gold data.
AUKBC-MULCRF overgenerates mentions, all other systems undergenerate mentions.
The average length of mentions predicted by
 AUKBC-MULCRF, DAggerCoref, LLM-LatticeNLP, and LLM-UWB is notably lower than in the gold data
 because they predicted single-word mentions only (or mostly in case of AUKBC-MULCRF).
All other systems have the distribution of mention lengths similar to the gold data,
although no system predicts long mentions (4 and 5+ words) more frequently than in the gold data.
}
\label{tab:stats-mentions-nonsingleton}
\end{table}

\begin{table}[H]\centering
\begin{tabular}{@{}l rrrr rrrrrr@{}}\toprule
                    & \MC{4}{    singleton mentions}     & \MC{6}{distribution of lengths}              \\\cmidrule(lr){2-5}\cmidrule(l){6-11}
system              &   total & per 1k & \MC{2}{length} &     0 &     1 &     2 &     3 &     4 &   5+ \\\cmidrule(lr){4-5}
                    &   count &  words &    max &  avg. &  [\%] &  [\%] &  [\%] &  [\%] &  [\%] & [\%] \\\midrule
gold                &  28,848 &    59 &    81 &  3.0 &  1.0 & 37.7 & 25.4 & 12.5 &  6.4 & 17.0 \\    
AUKBC-MULCRF        &       0 &     0 &     0 &  0.0 &  0.0 &  0.0 &  0.0 &  0.0 &  0.0 &  0.0 \\    
\baseline           &       0 &     0 &     0 &  0.0 &  0.0 &  0.0 &  0.0 &  0.0 &  0.0 &  0.0 \\    
\baselinegz         &       0 &     0 &     0 &  0.0 &  0.0 &  0.0 &  0.0 &  0.0 &  0.0 &  0.0 \\    
CorPipeEnsemble     &  30,061 &    62 &   133 &  2.9 &  1.2 & 36.9 & 26.9 & 12.6 &  6.3 & 16.2 \\    
CorPipeLarge        &  30,295 &    62 &   311 &  3.2 &  1.1 & 36.5 & 26.3 & 12.6 &  6.1 & 17.3 \\    
CorPipeXXL          &  29,858 &    61 &   133 &  2.9 &  1.2 & 37.1 & 26.7 & 12.6 &  6.1 & 16.3 \\    
DAggerCoref         &     315 &     1 &     1 &  0.9 &  7.3 & 92.7 &  0.0 &  0.0 &  0.0 &  0.0 \\    
LLM-Landcore        &  24,168 &    50 &    60 &  3.6 &  0.3 & 28.7 & 25.8 & 14.2 &  7.9 & 23.1 \\    
LLM-LatticeNLP      &  27,882 &    57 &     1 &  1.0 &  0.8 & 99.2 &  0.0 &  0.0 &  0.0 &  0.0 \\    
LLM-PortNLP         &  30,775 &    63 &    58 &  3.0 &  0.8 & 35.3 & 27.6 & 12.8 &  6.7 & 16.8 \\    
LLM-UWB             &  28,736 &    59 &     1 &  1.0 &  0.9 & 99.1 &  0.0 &  0.0 &  0.0 &  0.0 \\    
Stanza              &  23,075 &    47 &    58 &  2.9 &  1.1 & 39.5 & 24.1 & 12.4 &  6.4 & 16.4 \\    
\bottomrule\end{tabular}
\caption[]{Statistics on singleton mentions.
See the caption of Table~\ref{tab:stats-mentions-nonsingleton} for details.

\hskip5mm The two baseline systems and AUKBC-MULCRF do not attempt to predict singletons at all.
DAggerCoref heavily undergenerates singletons.
}
\label{tab:stats-mentions-singleton}
\end{table}

\begin{table}[H]\centering
\resizebox{\textwidth}{!}{
\begin{tabular}{@{}l @{\!}r@{~}r@{~}r @{~}r@{~}r@{~}r@{~}r@{~}r@{~}r@{~}r@{~}r@{~}r@{~}r@{}}\toprule
                    & \MC{3}{mention type [\%]}    & \MC{9}{distribution of head UPOS [\%]}      \\\cmidrule(lr){2-4}\cmidrule(l){5-14}
system              & w/empty & w/gap & non-tree
                                             &  NOUN &  PRON & PROPN &   DET &   ADJ &  VERB &   ADV &   NUM & \_~ & other \\\midrule
gold                &  10.7 &  0.6 &  1.7 & 35.3 & 36.8 & 14.9 &  5.7 &  1.2 &  1.6 &  1.9 &  0.4 &  1.5 &  0.7 \\    
AUKBC-MULCRF        &   7.1 &  0.0 &  1.1 & 42.4 & 43.6 & 11.6 &  0.0 &  0.0 &  0.0 &  2.4 &  0.0 &  0.0 &  0.0 \\    
\baseline           &  11.3 &  0.0 &  1.8 & 31.3 & 40.8 & 15.1 &  6.4 &  1.0 &  0.7 &  1.9 &  0.3 &  1.9 &  0.6 \\    
\baselinegz         &  11.5 &  0.0 &  1.8 & 31.2 & 41.0 & 15.1 &  6.4 &  1.0 &  0.7 &  1.9 &  0.3 &  1.7 &  0.6 \\    
CorPipeEnsemble     &  10.5 &  0.0 &  1.9 & 35.2 & 37.1 & 15.0 &  5.8 &  1.1 &  1.3 &  1.8 &  0.4 &  1.6 &  0.6 \\    
CorPipeLarge        &  10.2 &  0.0 &  2.3 & 35.4 & 37.0 & 14.9 &  5.8 &  1.1 &  1.3 &  1.8 &  0.4 &  1.7 &  0.6 \\    
CorPipeXXL          &  10.5 &  0.0 &  2.0 & 35.3 & 37.0 & 14.9 &  5.8 &  1.1 &  1.4 &  1.8 &  0.5 &  1.6 &  0.6 \\    
DAggerCoref         &   8.3 &  0.0 &  0.0 & 34.0 & 39.8 & 15.0 &  6.1 &  1.1 &  1.2 &  1.9 &  0.4 &  0.0 &  0.7 \\    
LLM-Landcore        &   6.8 &  0.0 &  2.3 & 37.3 & 28.1 & 17.6 &  5.0 &  1.1 &  1.0 &  1.9 &  0.4 &  6.6 &  1.0 \\    
LLM-LatticeNLP      &   9.6 &  0.0 &  0.0 & 34.0 & 30.6 & 15.1 &  5.9 &  1.0 &  1.1 &  1.7 &  0.4 &  9.6 &  0.6 \\    
LLM-PortNLP         &  10.3 &  0.0 &  1.8 & 33.8 & 31.2 & 14.7 &  6.0 &  1.0 &  1.0 &  1.6 &  0.4 &  9.7 &  0.6 \\    
LLM-UWB             &   9.5 &  0.0 &  0.0 & 35.5 & 29.0 & 15.3 &  5.6 &  1.1 &  1.2 &  1.7 &  0.4 &  9.5 &  0.6 \\    
Stanza              &  10.4 &  0.0 &  1.8 & 34.9 & 37.1 & 15.4 &  5.9 &  1.0 &  1.1 &  1.8 &  0.4 &  1.7 &  0.6 \\    
\bottomrule\end{tabular}
}
\caption[]{Detailed statistics on non-singleton mentions.
The left part of the table shows the percentage of:
 mentions with at least one empty node (w/empty);
 mentions with at least one gap, i.e.\ discontinuous mentions (w/gap);
 and non-treelet mentions, i.e.\ mentions not forming a connected subgraph (catena) in the dependency tree (non-tree).
Note that these three types of mentions may be overlapping.

\hskip5mm  We can see that none of the systems attempts to predict discontinuous mentions.
AUKBC-MULCRF, DAggerCoref, LLM-LatticeNLP, and LLM-UWB 
 have a notably lower percentage of non-treelet mention spans
 (1.1\% for AUKBC-MULCRF, 0\% for the others),
 but this is simply explained by their inability to generate (enough) multi-word mentions.

\hskip5mm The right part of the table shows the distribution of mentions
  based on the universal part-of-speech tag (UPOS) of the head word.
Note that this distribution has to be interpreted with the total number of non-singleton mentions predicted (as reported in Table~\ref{tab:stats-mentions-nonsingleton}) in mind.
For example, 41.0\% of non-singleton mentions predicted by \baselinegz{} are pronominal (head=PRON),
  while there are only 35.3\% of pronominal non-singleton mentions in the gold data.
However, \baselinegz{} predicts actually less pronominal non-singleton mentions
 ($64{,}583 \times 41.0\% \approx 26{,}479$) than in the gold data ($76{,}034 \times 35.3\% \approx 26{,}840$).

\hskip5mm Note that the same word may be assigned a different UPOS tag in the predicted and gold data
(in case of empty nodes or if the gold data includes manual annotation).
The empty UPOS tag (\_) is present only in the empty nodes
and none of the LLM systems attempts to predict the actual UPOS tag of empty nodes.
Unlike last year, the baseline predictor of empty nodes predicts also their UPOS,
so the unconstrained systems using this predictor predict
that 79\% of the empty nodes are pronouns and 21\% have an empty UPOS
(which is very close to the 79\% of pronouns and 19\% of empty UPOS in the gold data).
}
\label{tab:stats-details}
\end{table}

\clearpage
\section{Statistics of the submitted systems on the French LitBankFr test set}
\label{sec:stats-litbankfr}
\centering
\begin{tabular}{@{}l rrrrr rrrrr@{}}\toprule
                    & \MC{5}{entities}                          & \MC{5}{distribution of lengths}     \\\cmidrule(lr){2-6}\cmidrule(l){7-11}
system              &   total & per 1k & \MC{2}{length} & range &    1 &     2 &     3 &     4 &   5+ \\\cmidrule(lr){4-5}
                    &   count &  words &    max &  avg. & p95   & [\%] &  [\%] &  [\%] &  [\%] & [\%] \\\midrule
gold                &   1,218 &    29 &   934 &  5.0 & 4,435 & 69.2 & 13.7 &  3.9 &  2.8 & 10.4 \\    
AUKBC-MULCRF        &   1,413 &    34 &   654 &  5.2 & 1,215 &  0.0 & 55.7 & 19.1 &  7.9 & 17.3 \\    
\baseline           &     379 &     9 &   411 & 12.2 & 2,627 &  0.0 & 50.7 & 15.0 &  6.6 & 27.7 \\    
\baselinegz         &     379 &     9 &   411 & 12.2 & 2,627 &  0.0 & 50.7 & 15.0 &  6.6 & 27.7 \\    
CorPipeEnsemble     &   1,382 &    33 &   920 &  4.4 & 1,474 & 67.5 & 14.5 &  4.7 &  2.6 & 10.6 \\    
CorPipeLarge        &   1,437 &    35 &   919 &  4.3 & 1,291 & 66.9 & 14.8 &  5.6 &  2.7 & 10.0 \\    
CorPipeXXL          &   1,394 &    34 &   929 &  4.4 & 1,533 & 68.6 & 14.1 &  4.5 &  3.0 &  9.8 \\    
DAggerCoref         &     512 &    12 &   776 &  9.7 &   872 &  0.2 & 40.8 & 16.4 & 10.4 & 32.2 \\    
LLM-Landcore        &   1,205 &    29 &   831 &  3.7 & 3,575 & 68.1 & 13.9 &  6.6 &  3.4 &  8.0 \\    
LLM-LatticeNLP      &   1,201 &    29 &   974 &  5.0 & 2,246 & 64.9 & 15.8 &  5.4 &  2.6 & 11.3 \\    
LLM-PortNLP         &   5,556 &   134 &   914 &  2.1 & 6,729 & 75.8 & 15.6 &  3.9 &  1.5 &  3.3 \\    
LLM-UWB             &     886 &    21 &   819 &  6.7 & 9,370 & 38.5 & 24.6 &  9.6 &  6.8 & 20.5 \\    
Stanza              &     993 &    24 &   924 &  5.8 & 6,336 & 46.0 & 28.4 &  9.2 &  3.6 & 12.8 \\    
\bottomrule\end{tabular}

\bigskip
\begin{tabular}{@{}l rrrr rrrrrr@{}}\toprule
                    & \MC{4}{non-singleton mentions}     & \MC{6}{distribution of lengths}              \\\cmidrule(lr){2-5}\cmidrule(l){6-11}
system              &   total & per 1k & \MC{2}{length} &     0 &     1 &     2 &     3 &     4 &   5+ \\\cmidrule(lr){4-5}
                    &   count &  words &    max &  avg. &  [\%] &  [\%] &  [\%] &  [\%] &  [\%] & [\%] \\\midrule
gold                &   5,218 &   126 &    22 &  1.5 &  0.0 & 69.2 & 21.8 &  4.6 &  1.8 &  2.5 \\    
AUKBC-MULCRF        &   7,385 &   178 &     3 &  1.0 &  0.0 & 99.3 &  0.5 &  0.2 &  0.0 &  0.0 \\    
\baseline           &   4,623 &   112 &    12 &  1.3 &  0.0 & 76.0 & 19.5 &  3.2 &  0.7 &  0.6 \\    
\baselinegz         &   4,623 &   112 &    12 &  1.3 &  0.0 & 76.0 & 19.5 &  3.2 &  0.7 &  0.6 \\    
CorPipeEnsemble     &   5,175 &   125 &    52 &  1.5 &  0.0 & 69.3 & 21.5 &  4.6 &  2.0 &  2.6 \\    
CorPipeLarge        &   5,147 &   124 &   100 &  1.6 &  0.0 & 69.5 & 21.4 &  4.5 &  1.8 &  2.8 \\    
CorPipeXXL          &   5,186 &   125 &    23 &  1.5 &  0.0 & 69.2 & 21.3 &  4.6 &  2.0 &  2.8 \\    
DAggerCoref         &   4,963 &   120 &     1 &  1.0 &  0.0 &100.0 &  0.0 &  0.0 &  0.0 &  0.0 \\    
LLM-Landcore        &   3,587 &    87 &    34 &  1.6 &  0.0 & 64.6 & 25.0 &  4.9 &  2.1 &  3.3 \\    
LLM-LatticeNLP      &   5,169 &   125 &     1 &  1.0 &  0.0 &100.0 &  0.0 &  0.0 &  0.0 &  0.0 \\    
LLM-PortNLP         &   7,495 &   181 &    29 &  1.8 &  0.0 & 61.1 & 24.8 &  5.5 &  3.4 &  5.2 \\    
LLM-UWB             &   5,587 &   135 &     1 &  1.0 &  0.0 &100.0 &  0.0 &  0.0 &  0.0 &  0.0 \\    
Stanza              &   5,293 &   128 &    26 &  1.5 &  0.0 & 69.9 & 22.3 &  4.3 &  1.7 &  1.8 \\    
\bottomrule\end{tabular}

\bigskip
\begin{tabular}{@{}l rrrr rrrrrr@{}}\toprule
                    & \MC{4}{    singleton mentions}     & \MC{6}{distribution of lengths}              \\\cmidrule(lr){2-5}\cmidrule(l){6-11}
system              &   total & per 1k & \MC{2}{length} &     0 &     1 &     2 &     3 &     4 &   5+ \\\cmidrule(lr){4-5}
                    &   count &  words &    max &  avg. &  [\%] &  [\%] &  [\%] &  [\%] &  [\%] & [\%] \\\midrule
gold                &     843 &    20 &    25 &  2.9 &  0.0 & 17.7 & 37.6 & 19.5 & 10.9 & 14.4 \\    
AUKBC-MULCRF        &       0 &     0 &     0 &  0.0 &  0.0 &  0.0 &  0.0 &  0.0 &  0.0 &  0.0 \\    
\baseline           &       0 &     0 &     0 &  0.0 &  0.0 &  0.0 &  0.0 &  0.0 &  0.0 &  0.0 \\    
\baselinegz         &       0 &     0 &     0 &  0.0 &  0.0 &  0.0 &  0.0 &  0.0 &  0.0 &  0.0 \\    
CorPipeEnsemble     &     933 &    23 &    45 &  2.7 &  0.0 & 20.9 & 39.3 & 17.3 &  9.6 & 12.9 \\    
CorPipeLarge        &     962 &    23 &    88 &  3.5 &  0.0 & 20.4 & 39.6 & 16.2 &  8.9 & 14.9 \\    
CorPipeXXL          &     956 &    23 &    94 &  2.9 &  0.0 & 21.5 & 39.1 & 17.3 &  9.4 & 12.7 \\    
DAggerCoref         &       1 &     0 &     1 &  1.0 &  0.0 &100.0 &  0.0 &  0.0 &  0.0 &  0.0 \\    
LLM-Landcore        &     821 &    20 &    42 &  4.6 &  0.0 &  8.4 & 31.2 & 13.3 & 11.4 & 35.7 \\    
LLM-LatticeNLP      &     779 &    19 &     1 &  1.0 &  0.0 &100.0 &  0.0 &  0.0 &  0.0 &  0.0 \\    
LLM-PortNLP         &   4,214 &   102 &    39 &  3.3 &  0.0 & 13.1 & 40.0 & 14.8 & 10.3 & 21.9 \\    
LLM-UWB             &     341 &     8 &     1 &  1.0 &  0.0 &100.0 &  0.0 &  0.0 &  0.0 &  0.0 \\    
Stanza              &     457 &    11 &    21 &  3.0 &  0.0 & 22.3 & 34.4 & 11.2 & 14.9 & 17.3 \\    
\bottomrule\end{tabular}

\bigskip
\resizebox{\columnwidth}{!}{
\begin{tabular}{@{}l @{\!}r@{~}r@{~}r @{~}r@{~}r@{~}r@{~}r@{~}r@{~}r@{~}r@{~}r@{~}r@{~}r@{}}\toprule
                    & \MC{3}{mention type [\%]}    & \MC{9}{distribution of head UPOS [\%]}      \\\cmidrule(lr){2-4}\cmidrule(l){5-14}
system              & w/empty & w/gap & non-tree
                                             &  NOUN &  PRON & PROPN &   DET &   ADJ &  VERB &   ADV &   NUM & \_~ & other \\\midrule
gold                &   0.0 &  0.0 &  6.8 & 21.5 & 50.7 &  9.4 & 17.3 &  0.2 &  0.1 &  0.4 &  0.1 &  0.0 &  0.2 \\    
AUKBC-MULCRF        &   0.0 &  0.0 &  0.0 & 31.8 & 62.2 &  5.9 &  0.0 &  0.0 &  0.0 &  0.0 &  0.0 &  0.0 &  0.0 \\    
\baseline           &   0.0 &  0.0 &  6.9 & 15.7 & 56.8 &  7.3 & 19.9 &  0.0 &  0.0 &  0.1 &  0.1 &  0.0 &  0.1 \\    
\baselinegz         &   0.0 &  0.0 &  6.9 & 15.7 & 56.8 &  7.3 & 19.9 &  0.0 &  0.0 &  0.1 &  0.1 &  0.0 &  0.1 \\    
CorPipeEnsemble     &   0.0 &  0.0 &  6.7 & 21.3 & 51.0 &  9.2 & 17.7 &  0.2 &  0.0 &  0.4 &  0.1 &  0.0 &  0.1 \\    
CorPipeLarge        &   0.0 &  0.0 &  7.0 & 21.3 & 50.8 &  9.1 & 18.0 &  0.2 &  0.1 &  0.3 &  0.2 &  0.0 &  0.1 \\    
CorPipeXXL          &   0.0 &  0.0 &  6.8 & 21.4 & 51.0 &  9.2 & 17.5 &  0.2 &  0.0 &  0.4 &  0.2 &  0.0 &  0.1 \\    
DAggerCoref         &   0.0 &  0.0 &  0.0 & 19.0 & 53.3 &  8.3 & 18.5 &  0.1 &  0.2 &  0.3 &  0.1 &  0.0 &  0.2 \\    
LLM-Landcore        &   0.0 &  0.0 &  6.2 & 27.5 & 44.5 & 11.9 & 15.7 &  0.1 &  0.1 &  0.2 &  0.0 &  0.0 &  0.0 \\    
LLM-LatticeNLP      &   0.0 &  0.0 &  0.0 & 21.0 & 51.5 &  9.0 & 17.8 &  0.2 &  0.0 &  0.2 &  0.1 &  0.0 &  0.2 \\    
LLM-PortNLP         &   0.0 &  0.0 &  4.9 & 31.9 & 45.6 &  6.1 & 13.1 &  0.3 &  2.1 &  0.3 &  0.1 &  0.0 &  0.4 \\    
LLM-UWB             &   0.0 &  0.0 &  0.0 & 25.8 & 47.4 &  9.3 & 16.4 &  0.3 &  0.1 &  0.3 &  0.5 &  0.0 &  0.1 \\    
Stanza              &   0.0 &  0.0 &  6.2 & 22.1 & 50.2 &  9.5 & 17.9 &  0.1 &  0.0 &  0.1 &  0.1 &  0.0 &  0.0 \\    
\bottomrule\end{tabular}
}

\clearpage
\section{Statistics of the submitted systems on the Dutch OpenBoek test set}
\label{sec:stats-openboek}
\begin{tabular}{@{}l rrrrr rrrrr@{}}\toprule
                    & \MC{5}{entities}                          & \MC{5}{distribution of lengths}     \\\cmidrule(lr){2-6}\cmidrule(l){7-11}
system              &   total & per 1k & \MC{2}{length} & range &    1 &     2 &     3 &     4 &   5+ \\\cmidrule(lr){4-5}
                    &   count &  words &    max &  avg. & p95   & [\%] &  [\%] &  [\%] &  [\%] & [\%] \\\midrule
gold                &   1,611 &    74 &   479 &  3.0 & 4,183 & 70.3 & 16.0 &  6.1 &  1.7 &  6.0 \\    
AUKBC-MULCRF        &     690 &    32 &   217 &  4.7 & 1,051 &  0.0 & 53.8 & 18.0 &  6.8 & 21.4 \\    
\baseline           &     267 &    12 &   211 &  8.7 & 2,319 &  0.0 & 56.9 & 14.6 &  9.0 & 19.5 \\    
\baselinegz         &     267 &    12 &   211 &  8.7 & 2,319 &  0.0 & 56.9 & 14.6 &  9.0 & 19.5 \\    
CorPipeEnsemble     &   2,115 &    97 &   436 &  2.4 &   451 & 74.3 & 14.1 &  4.5 &  1.7 &  5.3 \\    
CorPipeLarge        &   2,136 &    98 &   489 &  2.4 &   530 & 74.6 & 13.2 &  5.1 &  1.8 &  5.2 \\    
CorPipeXXL          &   2,037 &    93 &   435 &  2.5 &   560 & 74.3 & 13.7 &  4.7 &  2.2 &  5.1 \\    
DAggerCoref         &     541 &    25 &   311 &  6.2 &   536 &  0.2 & 51.2 & 20.5 &  8.1 & 20.0 \\    
LLM-Landcore        &     433 &    20 &   446 &  4.1 & 5,798 & 67.9 & 15.7 &  4.8 &  1.8 &  9.7 \\    
LLM-LatticeNLP      &   1,909 &    88 &   502 &  2.6 &   502 & 75.1 & 12.9 &  5.1 &  1.7 &  5.1 \\    
LLM-PortNLP         &   1,992 &    91 &   446 &  2.4 &    99 & 78.4 & 12.3 &  4.2 &  1.3 &  3.8 \\    
LLM-UWB             &   2,069 &    95 &   392 &  2.4 &   476 & 75.1 & 13.5 &  4.7 &  1.8 &  4.9 \\    
Stanza              &     514 &    24 &   556 &  5.5 & 6,026 & 46.7 & 28.4 &  7.8 &  3.9 & 13.2 \\    
\bottomrule\end{tabular}

\bigskip
\begin{tabular}{@{}l rrrr rrrrrr@{}}\toprule
                    & \MC{4}{non-singleton mentions}     & \MC{6}{distribution of lengths}              \\\cmidrule(lr){2-5}\cmidrule(l){6-11}
system              &   total & per 1k & \MC{2}{length} &     0 &     1 &     2 &     3 &     4 &   5+ \\\cmidrule(lr){4-5}
                    &   count &  words &    max &  avg. &  [\%] &  [\%] &  [\%] &  [\%] &  [\%] & [\%] \\\midrule
gold                &   3,632 &   167 &    20 &  1.5 &  0.0 & 71.2 & 20.0 &  4.7 &  1.5 &  2.5 \\    
AUKBC-MULCRF        &   3,215 &   147 &     4 &  1.0 &  0.0 & 97.9 &  1.8 &  0.2 &  0.1 &  0.0 \\    
\baseline           &   2,328 &   107 &     8 &  1.2 &  0.0 & 89.7 &  7.3 &  1.8 &  0.5 &  0.6 \\    
\baselinegz         &   2,328 &   107 &     8 &  1.2 &  0.0 & 89.7 &  7.3 &  1.8 &  0.5 &  0.6 \\    
CorPipeEnsemble     &   3,466 &   159 &    24 &  1.4 &  0.0 & 73.1 & 18.8 &  4.8 &  1.3 &  2.0 \\    
CorPipeLarge        &   3,490 &   160 &    15 &  1.4 &  0.0 & 73.0 & 19.0 &  4.5 &  1.3 &  2.2 \\    
CorPipeXXL          &   3,493 &   160 &    12 &  1.4 &  0.0 & 72.9 & 18.9 &  4.7 &  1.3 &  2.2 \\    
DAggerCoref         &   3,363 &   154 &     1 &  1.0 &  0.0 &100.0 &  0.0 &  0.0 &  0.0 &  0.0 \\    
LLM-Landcore        &   1,500 &    69 &    12 &  1.4 &  0.0 & 78.5 & 13.7 &  4.1 &  1.2 &  2.5 \\    
LLM-LatticeNLP      &   3,598 &   165 &     1 &  1.0 &  0.0 &100.0 &  0.0 &  0.0 &  0.0 &  0.0 \\    
LLM-PortNLP         &   3,249 &   149 &    12 &  1.4 &  0.0 & 75.7 & 17.4 &  4.1 &  0.9 &  1.8 \\    
LLM-UWB             &   3,398 &   156 &     1 &  1.0 &  0.0 &100.0 &  0.0 &  0.0 &  0.0 &  0.0 \\    
Stanza              &   2,602 &   119 &    38 &  1.4 &  0.0 & 78.1 & 12.7 &  6.9 &  1.0 &  1.3 \\    
\bottomrule\end{tabular}

\bigskip
\begin{tabular}{@{}l rrrr rrrrrr@{}}\toprule
                    & \MC{4}{    singleton mentions}     & \MC{6}{distribution of lengths}              \\\cmidrule(lr){2-5}\cmidrule(l){6-11}
system              &   total & per 1k & \MC{2}{length} &     0 &     1 &     2 &     3 &     4 &   5+ \\\cmidrule(lr){4-5}
                    &   count &  words &    max &  avg. &  [\%] &  [\%] &  [\%] &  [\%] &  [\%] & [\%] \\\midrule
gold                &   1,133 &    52 &    21 &  2.7 &  0.0 & 14.9 & 50.0 & 15.4 &  5.8 & 13.9 \\    
AUKBC-MULCRF        &       0 &     0 &     0 &  0.0 &  0.0 &  0.0 &  0.0 &  0.0 &  0.0 &  0.0 \\    
\baseline           &       0 &     0 &     0 &  0.0 &  0.0 &  0.0 &  0.0 &  0.0 &  0.0 &  0.0 \\    
\baselinegz         &       0 &     0 &     0 &  0.0 &  0.0 &  0.0 &  0.0 &  0.0 &  0.0 &  0.0 \\    
CorPipeEnsemble     &   1,572 &    72 &    24 &  2.5 &  0.0 & 20.7 & 49.6 & 12.8 &  4.9 & 11.9 \\    
CorPipeLarge        &   1,594 &    73 &    36 &  2.6 &  0.0 & 22.0 & 48.4 & 12.7 &  5.3 & 11.7 \\    
CorPipeXXL          &   1,513 &    69 &    24 &  2.5 &  0.0 & 20.2 & 50.0 & 12.9 &  5.2 & 11.8 \\    
DAggerCoref         &       1 &     0 &     1 &  1.0 &  0.0 &100.0 &  0.0 &  0.0 &  0.0 &  0.0 \\    
LLM-Landcore        &     294 &    13 &    21 &  3.3 &  0.0 & 12.9 & 40.1 & 16.7 &  6.8 & 23.5 \\    
LLM-LatticeNLP      &   1,434 &    66 &     1 &  1.0 &  0.0 &100.0 &  0.0 &  0.0 &  0.0 &  0.0 \\    
LLM-PortNLP         &   1,561 &    72 &    22 &  2.6 &  0.0 & 16.9 & 52.7 & 12.7 &  5.3 & 12.4 \\    
LLM-UWB             &   1,553 &    71 &     1 &  1.0 &  0.0 &100.0 &  0.0 &  0.0 &  0.0 &  0.0 \\    
Stanza              &     240 &    11 &    11 &  3.0 &  0.0 & 11.7 & 30.4 & 37.1 &  5.4 & 15.4 \\    
\bottomrule\end{tabular}

\bigskip
\resizebox{\columnwidth}{!}{
\begin{tabular}{@{}l @{\!}r@{~}r@{~}r @{~}r@{~}r@{~}r@{~}r@{~}r@{~}r@{~}r@{~}r@{~}r@{~}r@{}}\toprule
                    & \MC{3}{mention type [\%]}    & \MC{9}{distribution of head UPOS [\%]}      \\\cmidrule(lr){2-4}\cmidrule(l){5-14}
system              & w/empty & w/gap & non-tree
                                             &  NOUN &  PRON & PROPN &   DET &   ADJ &  VERB &   ADV &   NUM & \_~ & other \\\midrule
gold                &   0.0 &  0.0 &  0.6 & 26.7 & 57.4 & 13.0 &  0.4 &  0.6 &  0.2 &  1.2 &  0.2 &  0.0 &  0.2 \\    
AUKBC-MULCRF        &   0.0 &  0.0 &  0.0 & 21.8 & 68.0 & 10.2 &  0.0 &  0.0 &  0.0 &  0.0 &  0.0 &  0.0 &  0.0 \\    
\baseline           &   0.0 &  0.0 &  1.1 &  8.6 & 81.9 &  8.7 &  0.0 &  0.2 &  0.1 &  0.1 &  0.1 &  0.0 &  0.3 \\    
\baselinegz         &   0.0 &  0.0 &  1.1 &  8.6 & 81.9 &  8.7 &  0.0 &  0.2 &  0.1 &  0.1 &  0.1 &  0.0 &  0.3 \\    
CorPipeEnsemble     &   0.0 &  0.0 &  0.4 & 25.5 & 60.2 & 13.0 &  0.2 &  0.4 &  0.2 &  0.1 &  0.2 &  0.0 &  0.2 \\    
CorPipeLarge        &   0.0 &  0.0 &  0.5 & 25.8 & 59.7 & 12.9 &  0.3 &  0.4 &  0.3 &  0.2 &  0.1 &  0.0 &  0.3 \\    
CorPipeXXL          &   0.0 &  0.0 &  0.5 & 26.0 & 59.8 & 13.0 &  0.3 &  0.4 &  0.2 &  0.1 &  0.2 &  0.0 &  0.1 \\    
DAggerCoref         &   0.0 &  0.0 &  0.0 & 24.2 & 62.9 & 11.7 &  0.1 &  0.2 &  0.2 &  0.3 &  0.1 &  0.0 &  0.2 \\    
LLM-Landcore        &   0.0 &  0.0 &  0.7 & 16.9 & 59.1 & 22.5 &  0.3 &  0.7 &  0.3 &  0.1 &  0.0 &  0.0 &  0.1 \\    
LLM-LatticeNLP      &   0.0 &  0.0 &  0.0 & 26.9 & 58.6 & 12.5 &  0.1 &  0.7 &  0.3 &  0.4 &  0.1 &  0.0 &  0.3 \\    
LLM-PortNLP         &   0.0 &  0.0 &  0.3 & 23.5 & 63.7 & 11.6 &  0.2 &  0.4 &  0.2 &  0.1 &  0.1 &  0.0 &  0.2 \\    
LLM-UWB             &   0.0 &  0.0 &  0.0 & 26.0 & 59.7 & 12.5 &  0.2 &  0.7 &  0.3 &  0.1 &  0.1 &  0.0 &  0.4 \\    
Stanza              &   0.0 &  0.0 &  0.2 & 15.5 & 67.8 & 15.9 &  0.1 &  0.3 &  0.2 &  0.0 &  0.0 &  0.0 &  0.2 \\    
\bottomrule\end{tabular}
}

\clearpage
\section{Evolution of Codabench Submissions}
\label{sec:codalab-evol}
\begin{figure*}[h]
    \centering
    \includegraphics[width=0.9\textwidth]{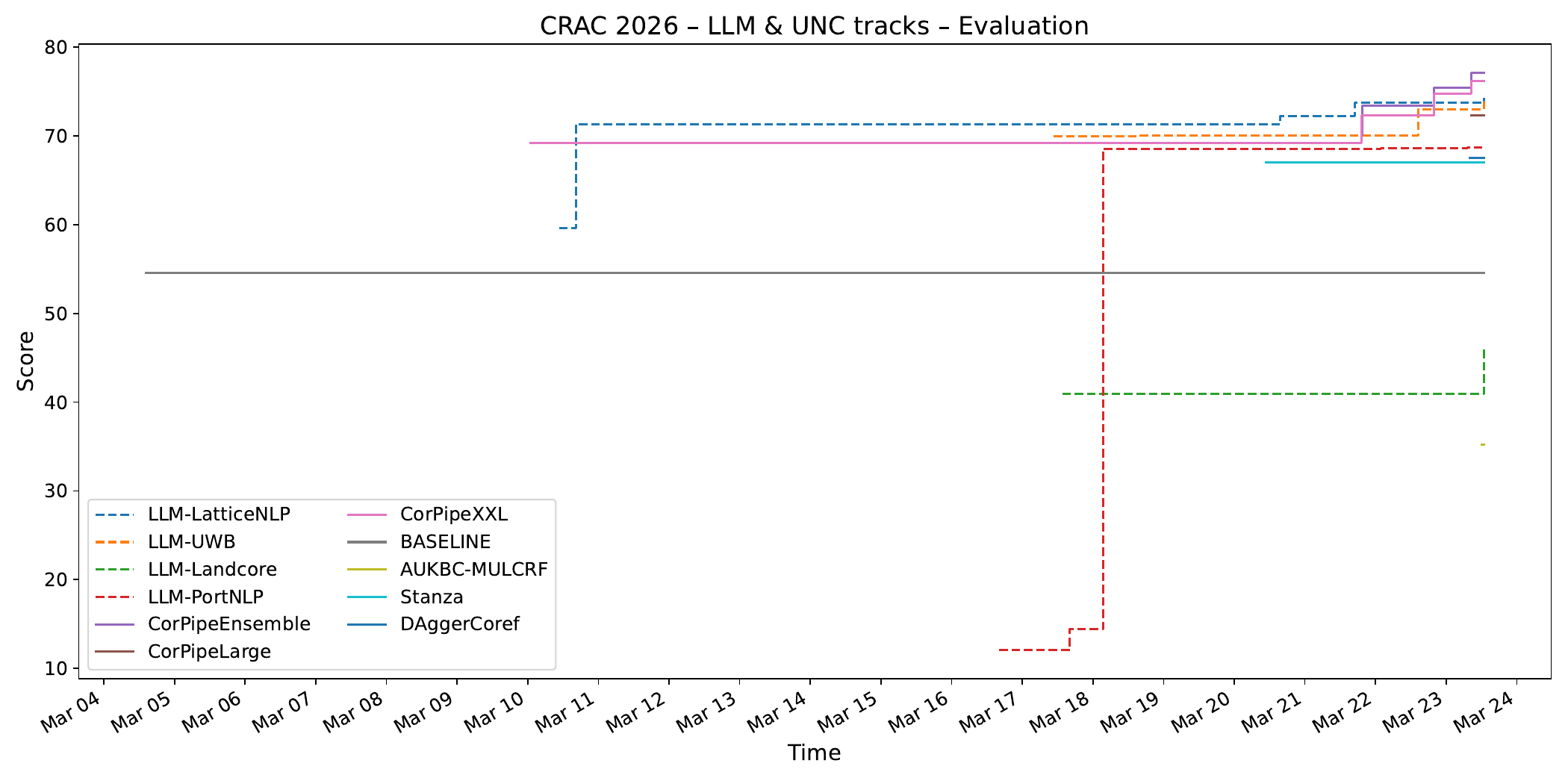}
    \caption{Evolution of Codabench Submissions in the %
    evaluation phase. The submissions to the LLM and Unconstrained track are shown by using the dashed and solid lines, respectively.}
    \label{fig:codalab-evol}
\end{figure*}

\end{document}